%% file: cdgp_neurips_2026.tex
\documentclass{article}

\PassOptionsToPackage{numbers,compress}{natbib}
\usepackage[preprint]{neurips_2026}
\setcitestyle{numbers,square,comma,sort&compress}

\usepackage[utf8]{inputenc}
\usepackage[T1]{fontenc}
\usepackage{hyperref}
\usepackage{url}
\usepackage{graphicx}
\usepackage{caption}
\usepackage{subcaption}
\usepackage{booktabs}
\usepackage{multirow}
\usepackage{makecell}
\usepackage{amsmath}
\usepackage{amssymb}
\usepackage{amsfonts}
\usepackage{placeins}
\usepackage{microtype}
\usepackage{xcolor}

\title{CDGP: Contrastive Dual Gaussian Processes for Weakly Supervised Anomaly Segmentation}
\input{authors}

\begin{document}

\maketitle

\begin{abstract}
Industrial visual inspection must both decide whether a product is defective and localize the defect, yet pixel-level masks are costly to collect at scale. Most anomaly-segmentation methods learn only from defect-free images and score deviations from normality. A true defect and an unusual-but-normal region, however, can both deviate substantially and receive similarly high scores. We propose Contrastive Dual Gaussian Processes (CDGP), a weakly supervised framework that models normal and anomaly inducing-variable predictive distributions over dense tokens. Its posterior-dominance statistic standardizes their predictive-mean difference by the joint predictive uncertainty, providing both spatial evidence and image-level confidence. This evidence complements hierarchical normal-reconstruction residuals for fine localization. All calibration uses training data only, without human pixel annotations or test-time fitting. Across MVTec AD~2, KSDD2, and VisA, CDGP ranks first among the evaluated methods on all MVTec AD~2 localization metrics and is first-place or competitive on KSDD2 and VisA. Factorized and matched linear-head controls delimit the contribution and scope of the linear-kernel Gaussian process (GP) formulation.
\end{abstract}

\input{intro}
\input{relatedwork}
\input{method}
\input{experiments}
\input{conclusion}

\bibliographystyle{plainnat}
\bibliography{references}

\clearpage
\appendix
\renewcommand{\thesection}{S\arabic{section}}
\renewcommand{\thetable}{S\arabic{table}}
\renewcommand{\thefigure}{S\arabic{figure}}
\renewcommand{\theequation}{S\arabic{equation}}
\setcounter{section}{0}
\setcounter{table}{0}
\setcounter{figure}{0}
\setcounter{equation}{0}

\input{supplement_content}

\end{document}

%% file: authors.tex
\author{%
  Seungjun Chu\thanks{Equal contribution.} \\
  Korea University
  \And
  Seokhee Han\footnotemark[1] \\
  Dartmouth College
  \AND
  Mateusz Nowak \\
  Dartmouth College
  \And
  Peter Chin \\
  Dartmouth College
}

%% file: intro.tex
\section{Introduction}

\begin{figure}[ht!]
    \centering

    \begin{subfigure}[t]{0.48\columnwidth}
        \centering
        \includegraphics[width=\linewidth]{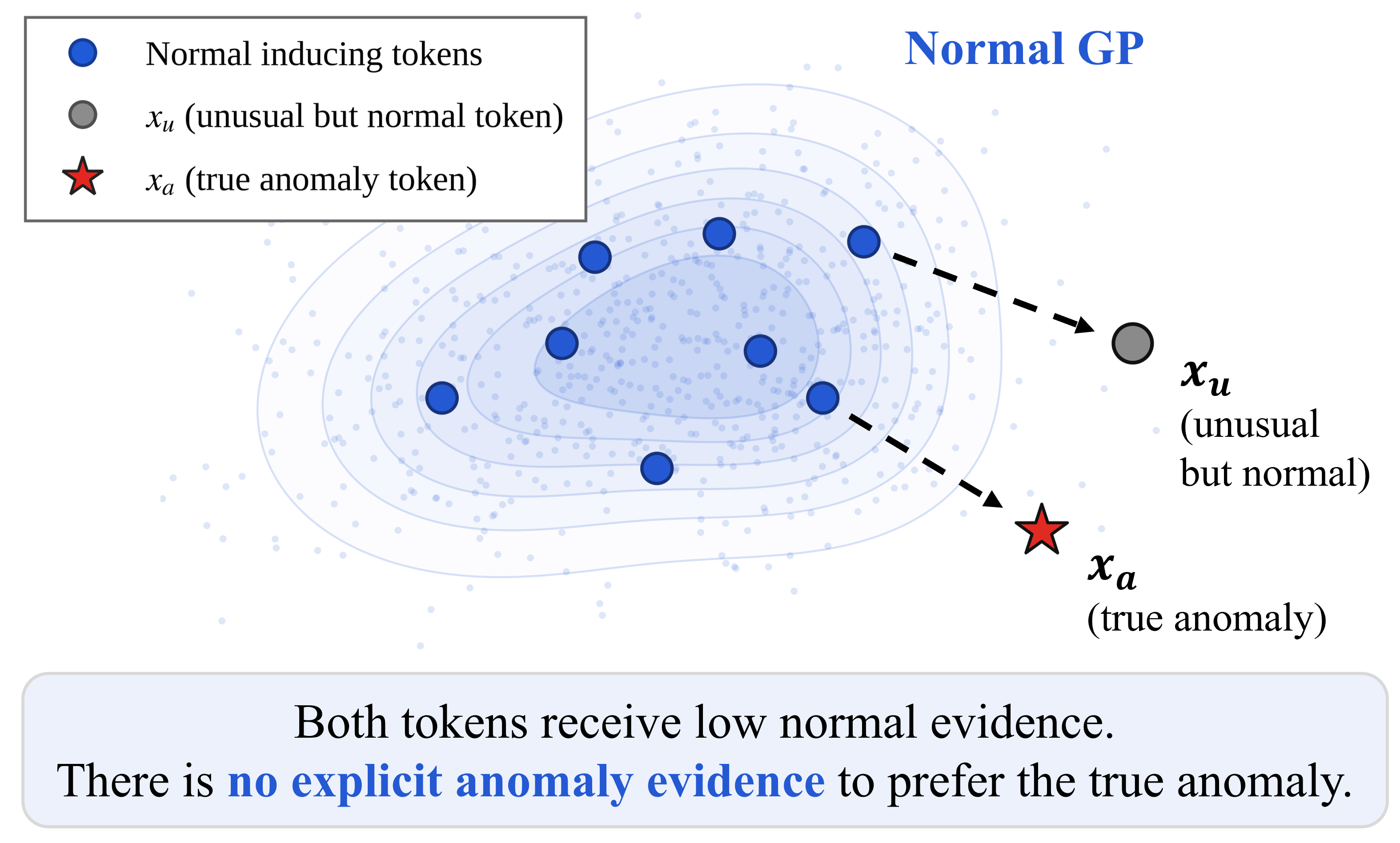}
        \caption{One-sided normality scoring.}
        \label{fig:cdgp_teaser_a}
    \end{subfigure}\hfill
    \begin{subfigure}[t]{0.48\columnwidth}
        \centering
        \includegraphics[width=\linewidth]{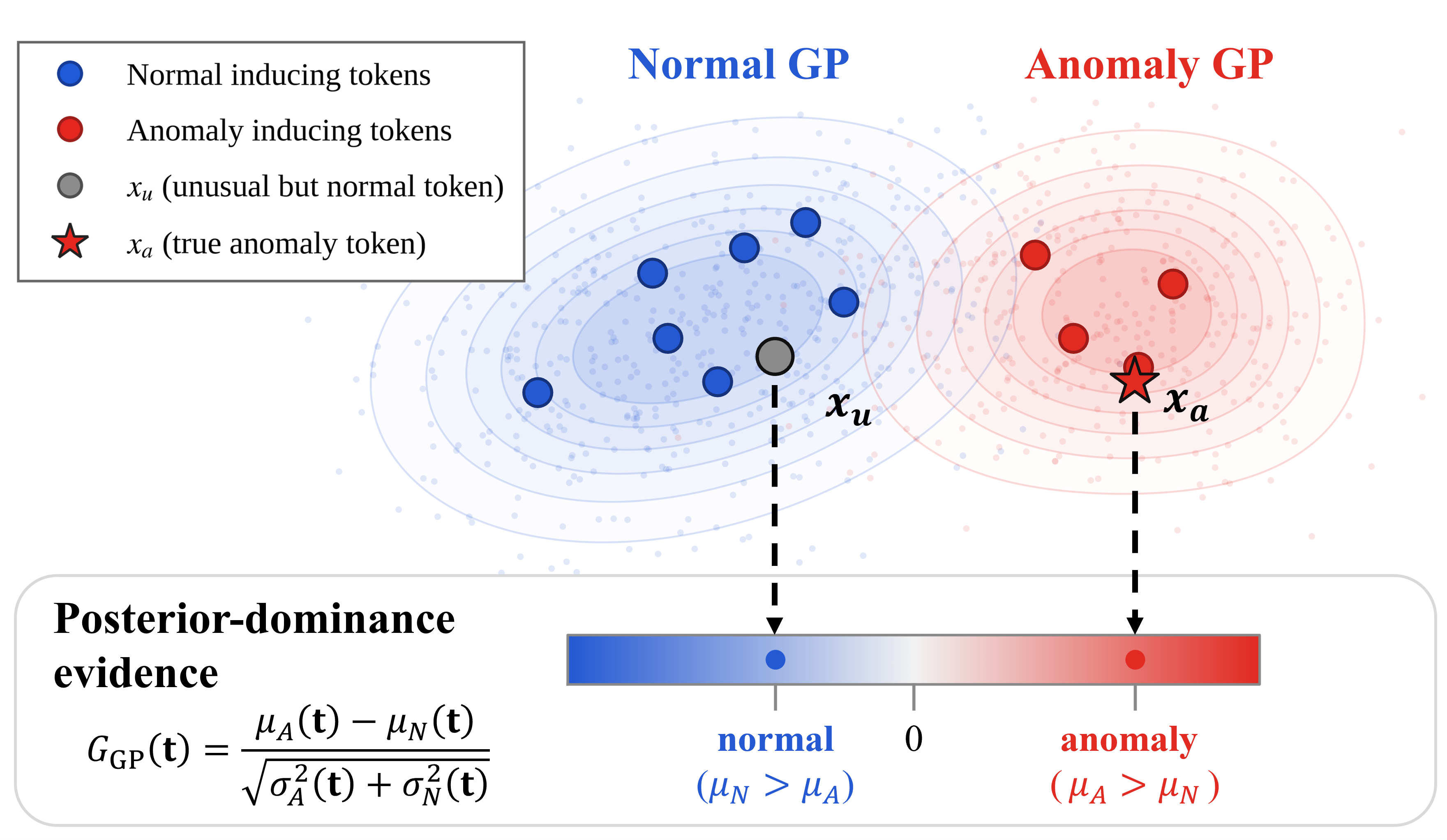}
        \caption{Contrastive dual predictive evidence in CDGP.}
        \label{fig:cdgp_teaser_b}
    \end{subfigure}

    \caption{
    \textbf{Conceptual motivation for CDGP.}
    One-sided normality scoring can assign equally low normal evidence to an unusual-but-normal region and a true anomaly.
    CDGP instead models normal and anomaly predictive distributions and scores their posterior dominance at each token.
    }
    \label{fig:cdgp_teaser}
\end{figure}

Industrial visual inspection must both decide whether a product is defective and localize the regions responsible for the failure. Accurate anomaly segmentation supports downstream tasks such as diagnosis, process control, and quality improvement. Yet collecting pixel-level defect masks is costly and time-consuming in real manufacturing environments~\cite{bergmann2019mvtec,zavrtanik2021draem}. Even when defective images are available, defects can be heterogeneous, spatially sparse, or visually ambiguous, making dense annotation slow and inconsistent~\cite{bergmann2019mvtec,hecklerkram2026mvtecad2,zou2022visa}. This annotation bottleneck has made industrial anomaly segmentation a natural testbed for learning with limited supervision.

Most existing industrial anomaly segmentation methods follow a normal-only learning paradigm, in which models are trained using defect-free images and anomalies are detected as deviations from normality. This paradigm has been highly successful because it avoids the need for pixel-level defect masks by leveraging the abundance of normal relative to defective samples. Representative methods model normal patch distributions, store nominal feature memories, reconstruct normal appearance, or distill normal representations for anomaly scoring~\cite{bergmann2019mvtec,defard2021padim,roth2022patchcore,zavrtanik2021draem,batzner2024efficientad}. Yet the normal-only assumption also hides an important practical requirement: the training set must be curated to exclude defective samples. Normal-only training is therefore not entirely free from supervision; it shifts the burden from pixel-level annotation to image-level screening or dataset curation.

In many inspection pipelines, the image-level normal/defective labels produced by such screening are naturally available, or at least cheaper to obtain than pixel-level masks. Once these labels exist, discarding defective images forfeits potentially useful information about how anomalies appear, even though exact pixel locations are unknown. This motivates weakly supervised anomaly segmentation, in which both normal and defective images, together with their image-level labels, are used during training without pixel-level annotations. Despite its practicality, weakly supervised anomaly segmentation has been studied less than normal-only anomaly detection on industrial benchmarks such as MVTec AD~2 and VisA~\cite{hecklerkram2026mvtecad2,zou2022visa}.

However, leveraging defective images with only image-level labels introduces a fundamental localization challenge. Such images typically contain small anomalous regions embedded in largely normal backgrounds, while their labels indicate only the presence of a defect, not its location. Under such weak supervision, activation maps extracted from an image-level classifier may highlight discriminative context that is sufficient for image-level recognition but misaligned with the actual defect regions~\cite{zhou2016cam,selvaraju2017gradcam}. This limitation is particularly severe in industrial anomaly segmentation, where defects do not form a coherent visual category and may vary widely in shape, texture, scale, and location. While weakly supervised semantic segmentation has developed increasingly sophisticated class-localization and pseudo-labeling strategies~\cite{xu2022mctformer,ru2023toco}, industrial anomaly segmentation requires localizing open-ended deviations rather than recurring semantic object categories.

This localization ambiguity calls for evidence defined at dense locations rather than localization derived only post hoc from an image classifier. We adopt a contrastive view: a token is anomalous when the anomaly predictive function is likely to exceed the normal function, not simply when a normality model explains it poorly. The resulting posterior-dominance statistic serves two coupled roles: it forms a spatial semantic map and, after pooling, estimates whether the image contains a defect. Multiscale normal-reconstruction residuals provide complementary fine spatial detail.

We propose \emph{Contrastive Dual Gaussian Processes} (CDGP), a weakly supervised anomaly-segmentation framework centered on sparse contrastive GP evidence. Figure~\ref{fig:cdgp_teaser} contrasts its dual predictive formulation with conventional one-sided normality scoring. CDGP represents each image as a grid of normalized multi-scale tokens and induces normal and anomaly predictive distributions from class-specific inducing sets. Sparse GPs provide compact kernel-based predictive functions over dense features~\cite{rasmussen2006gpml,snelson2006sparse,titsias2009variational,wilson2016dkl}. CDGP scores each token by posterior dominance between the two predictive distributions.

This direct comparison provides an explicit score at every dense location. To initialize the anomaly GP without masks, we form a noisy candidate pool from defective-image tokens least similar to selected normal tokens. It is only an initialization prior, not a pseudo-mask: an image-level objective subsequently learns both evidence functions despite normal background in defective images.

Token-level GP evidence alone can under-resolve thin defects. CDGP therefore adds a normal-only reverse student whose hierarchical reconstruction residuals supply fine-scale spatial precision. The Dual-GP dominance score supplies complementary spatial and image-level evidence. Normal calibration statistics standardize and fuse the two spatial branches, convert their score to normal-tail surprisal, and calibrate the pooled GP evidence at image level.

Our contributions are summarized as follows:
\begin{itemize}
\item We formulate weakly supervised industrial anomaly segmentation as posterior dominance between sparse normal and anomaly predictive functions over dense multi-scale tokens.

\item We use a mask-free, defect-enriched initialization and train the Dual-GP core with a minimal three-loss objective: max-margin MIL, covariance-normalized normal compactness, and covariance-normalized anomaly attraction.

\item We complement semantic GP contrast with hierarchical normal-reconstruction residuals and training-only calibration to obtain spatially precise predictions.

\item Under our matched weak-supervision split, CDGP ranks first among the evaluated methods on all MVTec AD~2 localization metrics and achieves first-place or competitive localization on KSDD2 and VisA; controls separate the complete Dual-GP pathway from modest variance-only refinement and linear-head alternatives.
\end{itemize}

%% file: relatedwork.tex
\section{Related Work}

\subsection{Normal-Only Industrial Anomaly Segmentation}

Industrial anomaly detection and segmentation have been extensively studied under the normal-only learning paradigm, in which models learn nominal appearance from anomaly-free images and identify defects as deviations from normal visual patterns. Representative methods differ in how they represent normality. DRAEM combines reconstruction with synthetically generated anomalies and discriminative segmentation training~\cite{zavrtanik2021draem}. PaDiM models pretrained patch embeddings at each spatial location with a multivariate Gaussian distribution~\cite{defard2021padim}, whereas PatchCore stores a representative coreset of nominal patch features and performs nearest-neighbor anomaly scoring~\cite{roth2022patchcore}. Distillation-based methods identify anomalies from discrepancies between teacher and student representations. This family includes reverse distillation~\cite{deng2022rd4ad} and EfficientAD, which emphasizes accurate and efficient inference~\cite{batzner2024efficientad}. These approaches avoid dense defect annotations but do not exploit defective training images when image-level normal/defective labels are available. CDGP instead learns its primary semantic evidence from both normal and defective image labels. It retains a normal-only reverse-student residual as complementary multiscale evidence, rather than treating deviation from normality as the sole anomaly signal.

\subsection{Weakly Supervised Localization and Industrial Defect Segmentation}

Weakly supervised semantic segmentation commonly learns dense localization from image-level class labels. A standard approach trains an image classifier, extracts class activation maps, and uses them directly or as pseudo labels for segmentation~\cite{zhou2016cam,selvaraju2017gradcam}. Transformer-based methods improve localization through class-specific tokens or token-level contrast. MCTformer learns localization maps using multiple class tokens~\cite{xu2022mctformer}, while ToCo contrasts patch and class tokens to improve weakly supervised segmentation~\cite{ru2023toco}. These methods generally assume recurring semantic categories with relatively stable visual structure. Industrial anomalies instead represent heterogeneous and open-ended deviations whose shape, appearance, scale, and location may vary substantially. Consequently, image-level classifiers can rely on discriminative context that supports classification without accurately localizing the underlying defect.

Industrial inspection studies have also explored forms of supervision weaker than dense pixel-level annotation. Bo{\v{z}}i{\v{c}} et al.\ study weak, mixed, and fully supervised surface-defect detection using annotations ranging from image-level labels to pixel-level masks~\cite{bozic2021mixed}. WeakREST supports image-level or optional coarse spatial supervision and combines block-wise classification with synthetic false anomalies~\cite{li2026weakrest}. SuperSimpleNet provides a unified framework for normal-only, weakly supervised, and fully supervised surface-defect detection; its weak mode learns from image tags and feature-space synthetic anomalies without human masks~\cite{rolih2024supersimplenet}. CDGP likewise uses only binary image labels, but differs by explicitly contrasting probabilistic normal and anomaly evidence and complementing it with normal-only hierarchical reconstruction.

\subsection{Probabilistic and Statistical Anomaly Scoring}

Industrial anomaly scoring often relies on statistical or distance-based comparisons to nominal features. PaDiM estimates spatial Gaussian distributions and uses Mahalanobis distance for anomaly scoring, whereas PatchCore measures distance to a representative memory of nominal patch features~\cite{defard2021padim,roth2022patchcore}. These approaches characterize abnormality primarily through deviation from normality.

GPs provide predictive distributions over latent functions through kernel-based modeling~\cite{rasmussen2006gpml}. Sparse GP approximations use a finite set of inducing points to reduce prediction cost~\cite{snelson2006sparse,titsias2009variational}. Deep kernel learning combines learned neural representations with kernel-based predictors~\cite{wilson2016dkl}. CDGP differs from one-sided statistical anomaly scoring by modeling both normal and anomalous evidence over a shared dense token space. It localizes defects through the standardized posterior dominance of two induced predictive functions, complements this semantic evidence with hierarchical normal-reconstruction residuals, and calibrates both signals using only held-out training data. Each trained model operates independently at test time without dense ground truth or test-time fitting.

%% file: method.tex
\section{Method}

\begin{figure*}[t]
    \centering
    \includegraphics[
        width=\textwidth,
        keepaspectratio
    ]{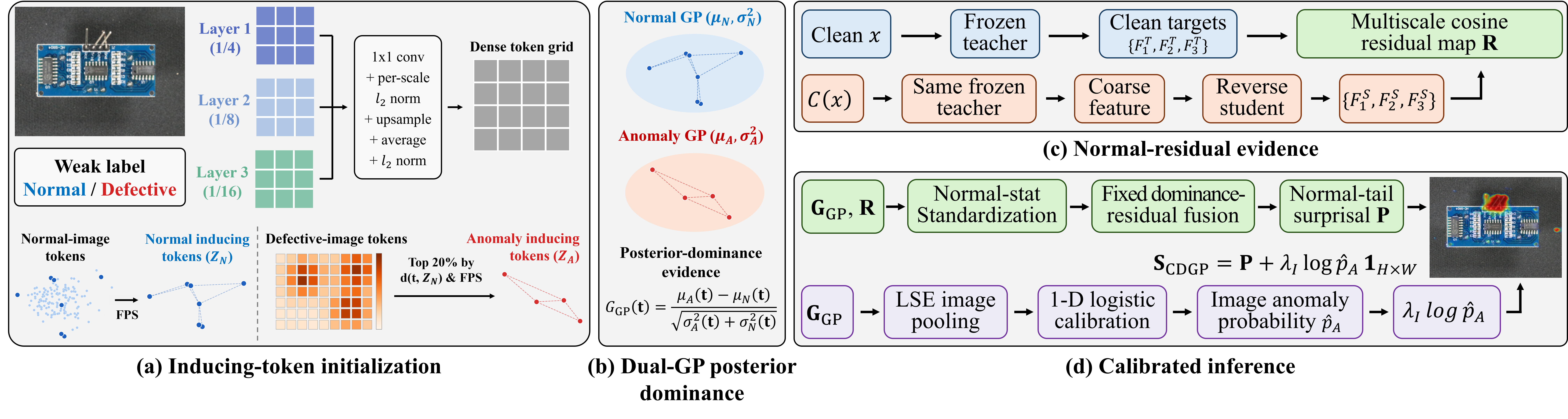}
    \caption{
    \textbf{Overview of CDGP.}
    (a) Multi-scale tokens initialize the normal and anomaly GP inducing sets through diverse normal sampling and hard-candidate mining.
    (b) Their joint predictive mean and variance define posterior-dominance evidence.
    (c) A frozen teacher and normal-only reverse student provide hierarchical residual evidence.
    (d) Training-only calibration fuses the two signals into the final anomaly map.
    }
    \label{fig:cdgp_overview}
\end{figure*}

Figure~\ref{fig:cdgp_overview} summarizes CDGP. Normal and anomaly inducing tokens define two sparse GP evidence functions whose posterior-dominance score provides dense semantic evidence and, after pooling, image-level confidence. After training the Dual GP, we freeze it, learn a complementary normal-residual branch, and combine both signals using training-only calibration.

\subsection{Dense Multi-Scale Token Representation}

We use an ImageNet-pretrained WideResNet-50-2 as the backbone encoder~\cite{zagoruyko2016wide}. Let $\mathbf{F}_1$, $\mathbf{F}_2$, and $\mathbf{F}_3$ denote feature maps from \texttt{layer1}, \texttt{layer2}, and \texttt{layer3}, respectively. We use these three stages to retain spatial detail while incorporating progressively higher-level representations.

Each feature map is projected to a shared latent dimension $D$ using a $1{\times}1$ convolution, normalized channel-wise, and aligned to the spatial resolution of the first stage:
\begin{equation}
\mathbf{T}
=
\ell_2
\left(
\frac{1}{3}
\sum_{\ell=1}^{3}
\operatorname{Up}_{\ell\rightarrow 1}
\left[
\ell_2\!\left(\phi_\ell(\mathbf{F}_\ell)\right)
\right]
\right)
\in
\mathbb{R}^{D\times H'\times W'},
\label{eq:token_fusion}
\end{equation}
where $\phi_\ell$ denotes the stage-specific projection, $\operatorname{Up}_{\ell\rightarrow 1}$ bilinearly resizes a feature map to the \texttt{layer1} resolution, and $H'=H/4$ and $W'=W/4$. For $\ell=1$, the resizing operator is the identity.

We denote the unit-norm token at spatial location $(h,w)$ by $\mathbf{t}_{hw}\in\mathbb{R}^{D}$. The resulting dense token grid is shared by the normal GP and the anomaly GP. The three backbone stages and the projection layers are optimized during Dual-GP training, while the earlier encoder stem remains frozen.

\subsection{Hard-Candidate Inducing Initialization}

Figure~\ref{fig:cdgp_overview}(a) summarizes the inducing-token initialization procedure. The normal GP and anomaly GP require different initialization strategies because normal images contain only nominal regions, whereas defective images contain both anomalous regions and substantial normal background.

For the normal GP, we collect dense tokens from normal training images and select a diverse subset using greedy farthest-point sampling, following the standard coreset construction used for patch representations~\cite{roth2022patchcore}. At each iteration, the procedure adds the candidate token whose maximum cosine similarity to the currently selected set is smallest. This produces a compact set of normal-GP inducing tokens $\mathbf{Z}_N$ that covers diverse regions of the nominal token space.

Directly sampling anomaly-GP inducing tokens from all tokens in defective images would allow normal background tokens to dominate the inducing set. We therefore compute the cosine distance of each defective-image token from the normal-GP inducing set:
\begin{equation}
d(\mathbf{t},\mathbf{Z}_N)
=
1-
\max_j
\mathbf{t}^{\top}\mathbf{Z}_{N,j}.
\label{eq:hard_distance}
\end{equation}
We retain the top $20\%$ of defective-image tokens with the largest $d(\mathbf{t},\mathbf{Z}_N)$ as the hard-candidate pool. The anomaly-GP inducing-token set $\mathbf{Z}_A$ is then selected from this pool using greedy farthest-point sampling. The pool is noisy initial support rather than a defect pseudo-mask or a set of pure defect prototypes; GP values and the encoder are subsequently learned from image labels.

The inducing-token locations remain fixed throughout training. The inducing-value distribution parameters, multi-scale projection layers, and trainable backbone stages are optimized using image-level supervision. We use $M_N=32$ and $M_A=16$ for every dataset and do not tune these budgets by category. Candidate selection uses only binary image labels and never accesses training masks.

\subsection{Contrastive Dual-GP Evidence}

\paragraph{Induced predictive functions.}
CDGP models normal and anomalous evidence as two sparse Gaussian-process functions over the shared token space. Sparse inducing-point approximations provide scalable kernel-based prediction from compact inducing sets \cite{rasmussen2006gpml,snelson2006sparse,titsias2009variational}.

For each evidence type $c\in\{N,A\}$, let $\mathbf{Z}_c\in\mathbb{R}^{M_c\times D}$ denote the inducing-token set and $\mathbf{u}_c\in\mathbb{R}^{M_c}$ the corresponding inducing function values. We parameterize an independent learned Gaussian distribution over each set of inducing values:
\begin{equation}
q(\mathbf{u}_c)
=
\mathcal{N}(\mathbf{m}_c,\mathbf{S}_c),
\qquad
\mathbf{S}_c=\mathbf{L}_c\mathbf{L}_c^\top.
\label{eq:inducing_distribution}
\end{equation}

Together with the sparse GP conditional, this distribution induces
\begin{equation}
q_c(f_c)
=
\int
p(f_c\mid\mathbf u_c,\mathbf Z_c)
q(\mathbf u_c)\,d\mathbf u_c .
\label{eq:induced_posterior_process}
\end{equation}
We call $q_c(f_c)$ the induced predictive process and use its first two moments. The shorthand ``posterior'' below refers operationally to this learned $q_c$, not to an exact Bayesian posterior.

We use a linear kernel over normalized tokens:
\begin{equation}
k(\mathbf{t},\mathbf{t}')
=
\mathbf{t}^{\top}\mathbf{t}'.
\label{eq:linear_kernel}
\end{equation}
Let
\[
\mathbf K_c
=
k(\mathbf Z_c,\mathbf Z_c)+\delta\mathbf I,
\]
where $\delta=10^{-3}$ is a fixed diagonal jitter used for numerical stability. We denote by $\mathbf k_{\mathbf t\mathbf Z_c}$ the vector of kernel evaluations between a query token $\mathbf t$ and the inducing set. The resulting predictive mean is
\begin{equation}
\mu_c(\mathbf{t})
=
\mathbf{k}_{\mathbf{t}\mathbf{Z}_c}
\mathbf{K}_c^{-1}
\mathbf{m}_c.
\label{eq:gp_mean}
\end{equation}

For the linear kernel, the corresponding predictive variance is
\begin{equation}
\sigma_c^2(\mathbf t)
=
\mathbf t^\top
\left[
\mathbf I
-\mathbf Z_c^\top\mathbf K_c^{-1}\mathbf Z_c
+\mathbf Z_c^\top\mathbf K_c^{-1}\mathbf S_c
 \mathbf K_c^{-1}\mathbf Z_c
\right]\mathbf t .
\label{eq:linear_gp_variance}
\end{equation}

\paragraph{Mean margin for weak training.}
The normal and anomaly GPs estimate evidence $\mu_N(\mathbf t)$ and $\mu_A(\mathbf t)$. Their predictive-mean margin is
\begin{equation}
s_{\mu}(\mathbf t)=\mu_A(\mathbf t)-\mu_N(\mathbf t).
\label{eq:mean_margin}
\end{equation}
This signed margin supplies the weak training signal in Section~\ref{sec:dual_gp_training}; it is not the final inference score.

\paragraph{Posterior-dominance evidence.}
Because the two inducing distributions are parameterized independently, their difference $\Delta f(\mathbf t)=f_A(\mathbf t)-f_N(\mathbf t)$ has induced marginal
\begin{equation}
q\!\left(\Delta f(\mathbf t)\right)
=\mathcal N\!\left(
s_{\mu}(\mathbf t),
\sigma_A^2(\mathbf t)+\sigma_N^2(\mathbf t)
\right).
\label{eq:contrastive_marginal}
\end{equation}
CDGP uses its standardized dominance statistic
\begin{equation}
\begin{aligned}
g(\mathbf t)
&=
\frac{\mu_A(\mathbf t)-\mu_N(\mathbf t)}
{\sqrt{\sigma_A^2(\mathbf t)+\sigma_N^2(\mathbf t)}},\\
\Pr_q[f_A(\mathbf t)>f_N(\mathbf t)]
&=\Phi(g(\mathbf t)).
\end{aligned}
\label{eq:cdgp_score}
\end{equation}
where $\Phi$ is the standard-normal cumulative distribution function (CDF); in implementation the summed variance is lower-bounded by $10^{-6}$. We use $g$, rather than applying $\Phi$, because they induce the same ordering while $g$ retains an unconstrained evidence scale. For dense prediction, $g_{hw}=g(\mathbf t_{hw})$.

\paragraph{Scope of the linear-kernel GP formulation.}
Because dense token targets are unavailable, $q(\mathbf u_c)$ is learned discriminatively using the image-level objective in Eq.~\ref{eq:total_loss}, rather than by optimizing a token-level likelihood or an evidence lower bound. The inducing locations $\mathbf Z_c$ are fixed after initialization. We therefore do not claim that $q_c(f_c)$ is the exact Bayesian posterior $p(f_c\mid\mathcal D)$ or that our training procedure is Titsias-style variational sparse-GP inference with jointly optimized inducing locations.

With the linear kernel, each predictive mean is linear in the learned token space:
\begin{equation}
\mu_c(\mathbf t)
=
\mathbb E_{q_c}[f_c(\mathbf t)]
=
\mathbf t^\top\boldsymbol\beta_c,
\qquad
\boldsymbol\beta_c
=
\mathbf Z_c^\top\mathbf K_c^{-1}\mathbf m_c .
\label{eq:linear_gp_interpretation}
\end{equation}
Consequently, the training margin is linear in the token:
\begin{equation}
s_{\mu}(\mathbf t)
=
\mathbf t^\top
(\boldsymbol\beta_A-\boldsymbol\beta_N).
\label{eq:linear_cdgp_score}
\end{equation}
We therefore do not attribute CDGP to nonlinear GP mean expressivity. Each $\boldsymbol\beta_c$ is restricted to the span of its class-specific inducing tokens. Unlike a homoscedastic linear head, however, the induced process ties this mean to a nonnegative, query-dependent predictive variance from the same $q(\mathbf u_c)$. Consequently, Eq.~\ref{eq:cdgp_score} is not generally a fixed linear head even though its numerator is linear. In the zero-jitter limit, the first variance component measures token mass outside the inducing span; the practical jitter yields a ridge-regularized analogue. The same induced covariance shapes auxiliary training and final inference. An engineered heteroscedastic linear model with an equivalent quadratic variance could reproduce this form. Our GP claim is therefore a compact inducing-variable construction of mean and uncertainty, not universal superiority over equivalent reparameterizations; the matched direct-linear control bounds its benefit against simpler mean scoring.

\subsection{Weakly Supervised Dual-GP Training}
\label{sec:dual_gp_training}

The Dual-GP core is trained end-to-end with three complementary image-supervised signals:
\begin{equation}
\mathcal{L}
=\mathcal L_{\rm MIL}
+\mathcal L_{\rm CMP}
+4\mathcal L_{\rm ABN}.
\label{eq:total_loss}
\end{equation}
These weights are fixed for every dataset. The three terms respectively separate normal and defective bags, reinforce normal-GP evidence on nominal tokens, and concentrate anomaly-GP evidence within defective images. No opposite-anchor repulsion or inducing-mean norm penalty is used in the model.

\paragraph{Max-margin multiple-instance learning.}
A defective image should contain at least one high-scoring token, whereas a normal image should contain no high-scoring token.
For image $i$, let
\begin{equation}
m_i
=
\max_{h,w}s_{\mu,i,hw}.
\label{eq:image_max_score}
\end{equation}
The corresponding loss is
\begin{equation}
\begin{split}
\mathcal{L}_{\mathrm{MIL}}^{(i)}
={} &
\mathbf{1}[y_i=1]\max(0,\gamma-m_i) \\
&+
\mathbf{1}[y_i=0]\max(0,m_i),
\end{split}
\label{eq:mil_loss}
\end{equation}
where $\gamma=0.5$ is fixed in all experiments.
The first term requires defective images to contain at least one token whose mean margin exceeds the threshold, while the second suppresses positive margins in normal images.

\paragraph{Covariance-normalized evidence losses.}
We use the predictive variance $\sigma_c^2(\mathbf{t})$ from the standard sparse-GP expression to define normalized evidence
\begin{equation}
\tilde{\mu}_c(\mathbf{t})
=
\frac{\mu_c(\mathbf{t})}
{\sqrt{\sigma_c^2(\mathbf{t})}+\epsilon},
\qquad c\in\{N,A\}.
\label{eq:normalized_evidence}
\end{equation}
This per-function normalization is used only in $\mathcal L_{\mathrm{CMP}}$ and $\mathcal L_{\mathrm{ABN}}$; final inference instead standardizes the difference distribution as in Eq.~\ref{eq:cdgp_score}.

Let $\mathcal I_N$ and $\mathcal I_A$ index normal and defective images, respectively, and let $j$ index spatial tokens. The normal compactness term is
\begin{equation}
\mathcal L_{\mathrm{CMP}}
=
-\mathbb E_{i\in\mathcal I_N,j}
\left[\widetilde\mu_N(\mathbf t_{ij})\right].
\label{eq:compact_loss}
\end{equation}
It encourages high normalized normal-GP evidence over nominal images.

For each defective image, we form a detached spatial attention map from the current mean-margin score:
\begin{equation}
a_{i,j}
=
\operatorname{softmax}_{j}
\left[
\operatorname{stopgrad}
\left(
\frac{s_{\mu,i,j}-\overline{s}_{\mu,i}}
{\operatorname{std}_{j}(s_{\mu,i,j})+\epsilon}
\right)
\right].
\label{eq:abnormal_attention}
\end{equation}
The anomaly-attraction loss then reinforces normalized anomaly-GP evidence at the attended locations:
\begin{equation}
\mathcal L_{\mathrm{ABN}}
=
-\mathbb E_{i\in\mathcal I_A}
\left[
\sum_j a_{i,j}\widetilde\mu_A(\mathbf t_{ij})
\right].
\label{eq:abnormal_attraction_loss}
\end{equation}
Unlike the MIL term, which acts on the most anomalous token in each bag, $\mathcal L_{\mathrm{CMP}}$ and $\mathcal L_{\mathrm{ABN}}$ shape the two class-specific evidence functions over dense locations. Their covariance normalization supplies a covariance-aware training signal, while Eq.~\ref{eq:cdgp_score} uses the joint predictive variance directly at inference.

\subsection{Hierarchical Normal-Residual Evidence}
\label{sec:hierarchical_residual}

The Dual GP provides semantic normal-versus-anomaly evidence, but its token grid can under-resolve thin defects. We therefore complement it with a normal-only reconstruction signal at three feature scales, following the reverse-distillation principle~\cite{deng2022rd4ad}. A frozen ImageNet-pretrained teacher produces features $\{\mathbf{F}^{T}_{\ell}(\mathbf{x})\}_{\ell=1}^{3}$. A reverse student takes the teacher's coarse representation of a corrupted normal image $\widetilde{\mathbf{x}}$ and reconstructs the clean teacher hierarchy:
\begin{equation}
 \{\mathbf{F}^{S}_{\ell}(\widetilde{\mathbf{x}})\}_{\ell=1}^{3}
 =
 \mathcal{S}_{\psi}
 \left(\mathbf{F}^{T}_{3}(\widetilde{\mathbf{x}})\right).
 \label{eq:reverse_student}
\end{equation}
The corruption combines weak additive noise, local averaging, and random erasing; its detailed probabilities are fixed across datasets and reported in the supplement. Only normal images from the second-stage fit subset are used to optimize the student. At scale $\ell$, the cosine reconstruction residual is
\begin{equation}
 r_{\ell,hw}(\mathbf{x})
 =
 1-
 \left\langle
 \overline{\mathbf{F}}^{T}_{\ell,hw}(\mathbf{x}),
 \overline{\mathbf{F}}^{S}_{\ell,hw}(\mathbf{x})
 \right\rangle,
 \label{eq:hierarchical_residual}
\end{equation}
where the overline denotes channel-wise $\ell_2$ normalization. The student is optimized using the mean reconstruction residual together with a hard-tail term over the largest residuals and a smooth feature-reconstruction term:
\begin{equation}
\begin{split}
 \mathcal{L}_{\mathrm{rec}}
 =
 \frac{1}{3}\sum_{\ell=1}^{3}
 \left[
 \operatorname{Mean}(r_{\ell})
 +\lambda_h\operatorname{TopMean}_{q}(r_{\ell}) \right.\\
 \left.
 +\lambda_s\mathcal{L}_{\mathrm{smooth}}^{(\ell)}
 \right].
\end{split}
 \label{eq:student_loss}
\end{equation}
We set $q=0.10$, $\lambda_h=0.65$, and $\lambda_s=0.10$, where $\mathcal L_{\mathrm{smooth}}^{(\ell)}$ is the Smooth-$L_1$ distance between the channel-normalized teacher and student features. At inference, corruption is disabled: the same clean test image is passed to the teacher, and the student reconstructs its hierarchy from the teacher's clean coarse feature. After alignment to the Dual-GP grid, the default hierarchical residual map is the average of the three scale-wise residuals,
\begin{equation}
 \mathbf{R}(\mathbf{x})
 =
 \frac{1}{3}\sum_{\ell=1}^{3}
 \operatorname{Up}_{\ell\rightarrow 1}
 \left(\mathbf{r}_{\ell}(\mathbf{x})\right).
 \label{eq:residual_map}
\end{equation}

\subsection{Calibrated Inference}
\label{sec:calibrated_inference}

For the second-stage student and calibration procedure, we partition the weak training set into a fit subset and a disjoint $20\%$ calibration subset. The Dual GP is frozen before this stage. Normal calibration images are used for branch standardization and spatial tail estimation; binary labels from both normal and defective calibration images are used only for the image-level calibrator described below.

We evaluate Eq.~\ref{eq:cdgp_score} over the token grid to obtain the posterior-dominance map
\begin{equation}
 \mathbf{G}_{\rm GP}(\mathbf{x})
 =
 \{g(\mathbf{t}_{hw})\}_{h,w}.
 \label{eq:gp_map}
\end{equation}
For each spatial branch $b\in\{G_{\rm GP},R\}$, scalar location and scale statistics $(m_b,s_b)$ are the mean and standard deviation estimated from normal calibration images, and $\mathcal{Z}_b(\mathbf{B})=(\mathbf{B}-m_b)/(s_b+\epsilon)$. The fused spatial evidence is
\begin{equation}
 \mathbf{U}(\mathbf{x})
 =
 w\,\mathcal{Z}_{G_{\rm GP}}(\mathbf{G}_{\rm GP})
 +(1-w)\,\mathcal{Z}_R(\mathbf{R}),
 \qquad w=0.025.
 \label{eq:default_fusion}
\end{equation}
Both maps are bilinearly upsampled to the input resolution before fusion.

Let $\{u^{(N)}_{d,n}\}_{n=1}^{N_d}$ be fused pixel scores collected from normal calibration images of dataset category $d$. For a query score $u$, we estimate its finite-sample upper-tail probability by
\begin{equation}
\widehat p_{N,d}(u)
=
\frac{
1+\sum_{n=1}^{N_d}\mathbf 1[u^{(N)}_{d,n}>u]
}{N_d+1}.
\label{eq:normal_tail_probability}
\end{equation}
The calibrated spatial evidence is the corresponding normal-tail surprisal,
\begin{equation}
P_{hw}(\mathbf x)
=
-\log\max\!\left\{
\widehat p_{N,d}(U_{hw}(\mathbf x)),e^{-16}
\right\}.
\label{eq:tail_surprisal}
\end{equation}
For image-level detection, the dominance scores are aggregated using smooth log-sum-exp pooling with fixed $r=8$:
\begin{equation}
 a(\mathbf{x})
 =
 \frac{1}{r}
 \log\!\left(
 \frac{1}{H'W'}\sum_{h,w}\exp(r g_{hw})
 \right).
 \label{eq:image_score}
\end{equation}
A one-dimensional logistic calibrator maps $a(\mathbf{x})$ to an anomaly probability $\widehat{p}_A(\mathbf{x})$. Before fitting, image scores are standardized using the median and IQR scale of normal calibration scores. We constrain the fitted slope to be non-negative so that calibration cannot reverse the predefined anomaly-score direction. The final CDGP score is
\begin{equation}
 \mathbf{S}_{\mathrm{CDGP}}(\mathbf{x})
=
\mathbf{P}(\mathbf{x})
+
\lambda_I\log\widehat{p}_A(\mathbf{x})\,
\mathbf{1}_{H\times W},
\qquad
\lambda_I=1/3.
 \label{eq:cdgp_final_score}
\end{equation}
Here the scalar $\log\widehat p_A(\mathbf x)$ is broadcast uniformly over the $H\!\times\!W$ map. It therefore cannot change pixel ordering within an image; it suppresses all locations in images judged likely normal and changes only cross-image score comparisons. We use a fixed moderate $\lambda_I$ so this global confidence remains subordinate to spatial evidence; the supplement reports its sensitivity. The image-level anomaly score is $\widehat{p}_A(\mathbf{x})$. All calibration quantities are estimated independently for each trained model using only its training partition.

%% file: experiments.tex
\section{Experiments}

\input{tables/main_v2}

\subsection{Experimental Setup}

\paragraph{Supervision protocol.}
CDGP uses only normal/defective image labels; spatial annotations are excluded from training, tuning, calibration, and checkpoint selection. Masks are loaded only for final evaluation of the fixed-budget last checkpoint. Here ``weak'' denotes label granularity, not a low-shot claim; all primary experiments use the declared matched or high-shot splits.

\paragraph{Evaluation metrics.}
We report image AUROC (AUROC-I), pixel AUROC (AUROC-P), and AUPRO at $\mathrm{FPR}_{\max}=0.3$, plus AUPRO@0.05 for MVTec AD~2 \cite{hecklerkram2026mvtecad2}. Mean$\pm$std uses three seeds; multi-category datasets are averaged by category within seed and then across seeds.

\subsection{Datasets}

\paragraph{MVTec AD 2.}
MVTec AD 2 is our primary benchmark for comparisons and ablations~\cite{hecklerkram2026mvtecad2}. Because its official training and validation sets are defect-free, we follow WeakREST~\cite{li2026weakrest} and construct category-wise stratified 80/20 splits from the publicly labeled normal and defective images. Each seed-specific disjoint split is shared by CDGP and all rerun baselines.

\paragraph{KSDD2.}
KSDD2 contains normal and defective images from an industrial surface inspection system~\cite{bozic2021mixed}. We use its official high-shot train/test split and reduce each training annotation to a binary image-level label.

\paragraph{VisA.}
VisA comprises twelve industrial categories with diverse anomalies~\cite{zou2022visa}. We use its official 2cls\_highshot train/test split, which provides both normal and anomalous training images, using only image-level labels for
supervision.

\subsection{Comparison with Representative Baselines}

We compare five normal-only methods~\cite{defard2021padim,roth2022patchcore, deng2022rd4ad,zavrtanik2021draem,batzner2024efficientad} with image-level Grad-CAM, MIL-TopK, SuperSimpleNet, WeakREST, and Static max \cite{selvaraju2017gradcam,rolih2024supersimplenet,li2026weakrest}. Static max uses our initialization but direct token--anchor similarity. Reruns share manifests and evaluation; normal-only methods are complementary references.

Under the matched MVTec protocol, Table~\ref{tab:combined_comparison} ranks CDGP first on all three localization metrics, ahead by $4.8/12.7/11.9$ points. Within image-level methods, it ranks first on both VisA metrics and first/second on KSDD2. Across both supervision groups, it ranks among the top two on all KSDD2 and VisA localization metrics.

\subsection{Component Analysis on MVTec AD 2}
\label{sec:ablation}

With matched manifests and calibration, Table~\ref{tab:component_analysis} factorizes dense posterior dominance $G_{\rm GP}$ and its image calibration $\widehat p_A$. From the GP-free residual pipeline, spatial evidence alone adds $9.8/25.7/17.8$, image evidence alone adds $11.0/29.4/18.4$, and both add $11.4/31.1/21.0$ AUROC-P/AUPRO@.3/AUPRO@.05 points. Because both paths share the same statistic, gains need not add. In particular, $w=0$ retains $\widehat p_A$; hence the full-minus-image difference ($+0.4/+1.7/+2.6$) is a conditional spatial refinement given GP image calibration, not the end-to-end effect of removing the Dual GP.

\input{tables/component_analysis}

Changing only same-checkpoint mean-margin scoring to posterior dominance improves the three MVTec AD~2 localization metrics by $0.1/0.3/0.3$ points. This controlled comparison isolates predictive variance at inference and shows that posterior standardization acts as a modest refinement rather than the primary source of the full-pipeline gain. A matched direct-linear control is reported in the supplement; its comparable performance bounds our claim to a coherent inducing-variable mean--variance construction rather than superior mean expressivity.

\begin{figure}[!t]
    \centering
    \includegraphics[width=1\columnwidth]{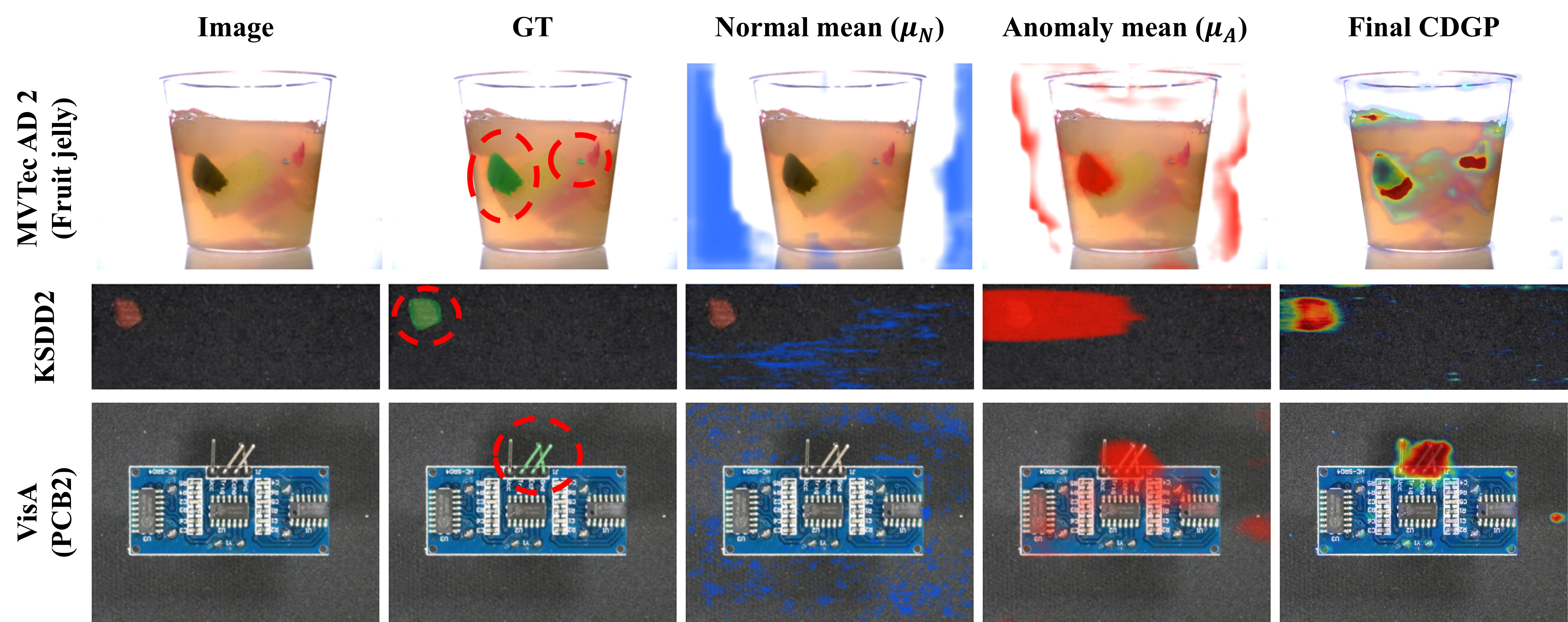}
    \caption{Input, ground truth, normal/anomaly predictive evidence, and final CDGP map.}
    \label{fig:qualitative_results}
\end{figure}

Figure~\ref{fig:qualitative_results} shows GP responses and final localization across all three datasets.

%% file: tables/main_v2.tex
\begin{table*}[t]
\centering
\footnotesize
\newcommand{\ms}[2]{$#1_{\pm#2}$}
\newcommand{\msb}[2]{$\mathbf{#1}_{\pm#2}$}
\newcommand{\msu}[2]{\underline{$#1$}$_{\pm#2}$}
\setlength{\tabcolsep}{1pt}

\begin{tabular*}{\textwidth}{@{\extracolsep{\fill}}lccccccc@{}}
\toprule
\multirow{2}{*}{Method}
& \multicolumn{3}{c}{MVTec AD~2}
& \multicolumn{2}{c}{KSDD2}
& \multicolumn{2}{c}{VisA} \\
\cmidrule(lr){2-4}\cmidrule(lr){5-6}\cmidrule(lr){7-8}
& AUROC-P & AUPRO@.3 & AUPRO@.05
& AUROC-P & AUPRO@.3
& AUROC-P & AUPRO@.3 \\
\midrule
\multicolumn{8}{l}{\textit{Normal-only supervision}} \\
PaDiM~{\scriptsize \cite{defard2021padim}} & \ms{85.0}{1.2} & \ms{49.5}{1.1} & \ms{18.2}{1.3} & \ms{93.4}{0.2} & \ms{87.1}{1.2} & \ms{97.2}{0.2} & \ms{83.3}{1.4} \\
PatchCore~{\scriptsize \cite{roth2022patchcore}} & \ms{91.2}{1.2} & \ms{63.0}{1.3} & \ms{36.1}{2.3} & \ms{91.6}{0.0} & \ms{85.2}{0.1} & \ms{97.6}{0.0} & \ms{85.5}{0.1} \\
RD4AD~{\scriptsize \cite{deng2022rd4ad}} & \ms{86.7}{0.9} & \ms{59.2}{0.6} & \ms{29.6}{1.6} & \ms{95.0}{0.1} & \ms{90.0}{0.1} & \msu{99.0}{0.0} & \ms{93.0}{0.1} \\
DRAEM~{\scriptsize \cite{zavrtanik2021draem}} & \ms{57.3}{4.0} & \ms{23.9}{5.4} & \ms{7.1}{0.9} & \ms{51.5}{7.4} & \ms{26.7}{6.2} & \ms{53.8}{4.8} & \ms{31.1}{5.0} \\
EfficientAD~{\scriptsize \cite{batzner2024efficientad}} & \ms{86.5}{2.4} & \ms{53.6}{0.4} & \msu{36.8}{0.5} & \ms{91.2}{0.5} & \ms{81.2}{1.2} & \msb{99.4}{0.0} & \msb{96.9}{0.1} \\
\midrule
\multicolumn{8}{l}{\textit{Image-level supervision}} \\
Classifier + Grad-CAM & \msu{92.0}{0.7} & \msu{71.7}{1.9} & \ms{27.9}{3.4} & \ms{93.1}{0.4} & \ms{87.3}{1.1} & \ms{91.8}{0.8} & \ms{79.5}{0.9} \\
MIL-TopK & \ms{55.6}{2.3} & \ms{19.5}{5.1} & \ms{6.9}{1.4} & \ms{91.6}{0.2} & \ms{87.1}{0.4} & \ms{47.1}{2.0} & \ms{22.2}{0.9} \\
SuperSimpleNet~(weak)~{\scriptsize \cite{rolih2024supersimplenet}} & \ms{85.0}{1.0} & \ms{48.1}{3.9} & \ms{26.2}{3.7} & \ms{93.2}{1.0} & \msb{95.1}{0.5} & \ms{89.9}{3.4} & \ms{80.1}{5.7} \\
WeakREST~{\scriptsize \cite{li2026weakrest}} & \ms{84.4}{1.2} & \ms{54.5}{0.6} & \ms{28.5}{2.3} & \ms{94.3}{3.3} & \ms{88.0}{4.0} & \ms{96.6}{2.0} & \ms{86.2}{2.1} \\
Static max similarity & \ms{90.3}{1.1} & \ms{69.1}{1.9} & \ms{10.4}{1.7} & \msu{97.1}{0.5} & \ms{92.8}{0.8} & \ms{85.5}{1.7} & \ms{65.0}{4.3} \\
\midrule
\textbf{CDGP (Ours)} & \msb{96.8}{0.7} & \msb{84.4}{0.7} & \msb{48.7}{1.8} & \msb{98.0}{0.6} & \msu{93.5}{0.1} & \msb{99.4}{0.1} & \msu{96.1}{0.7} \\
\bottomrule
\end{tabular*}
\par\vspace{0.5mm}
\textbf{(a) Localization performance}

\par\vspace{2mm}
\begin{tabular*}{0.92\textwidth}{@{\extracolsep{\fill}}lccc@{}}
\toprule
Method & MVTec AD~2 & KSDD2 & VisA \\
& AUROC-I & AUROC-I & AUROC-I \\
\midrule
\multicolumn{4}{l}{\textit{Normal-only supervision}} \\
PaDiM & \ms{69.9}{1.4} & \ms{86.7}{1.3} & \ms{80.8}{1.8} \\
PatchCore & \ms{82.8}{1.5} & \ms{79.6}{0.8} & \ms{86.6}{0.4} \\
RD4AD & \ms{74.3}{2.4} & \ms{90.4}{1.4} & \ms{95.4}{0.2} \\
DRAEM & \ms{54.2}{5.3} & \ms{64.5}{11.7} & \ms{56.2}{2.4} \\
EfficientAD & \ms{85.6}{0.7} & \ms{95.6}{0.6} & \ms{97.7}{0.0} \\
\midrule
\multicolumn{4}{l}{\textit{Image-level supervision}} \\
Classifier + Grad-CAM & \msb{100.0}{0.0} & \ms{90.9}{4.8} & \ms{97.8}{0.7} \\
MIL-TopK & \msu{99.2}{0.8} & \ms{95.3}{0.4} & \ms{91.7}{0.7} \\
SuperSimpleNet~(weak) & \ms{97.5}{0.6} & \msb{99.5}{0.1} & \msb{99.3}{0.1} \\
WeakREST & \ms{78.1}{1.1} & \ms{91.6}{4.2} & \ms{88.6}{1.2} \\
Static max similarity & \msb{100.0}{0.0} & \ms{95.9}{0.7} & \ms{97.9}{0.2} \\
\midrule
\textbf{CDGP (Ours)} & \msb{100.0}{0.0} & \msu{97.4}{0.2} & \msu{98.4}{0.3} \\
\bottomrule
\end{tabular*}
\par\vspace{0.5mm}
\textbf{(b) Image-level detection performance}

\caption{\textbf{Comparison with representative industrial anomaly segmentation baselines on MVTec AD~2, KSDD2, and VisA.} Mean\,$\pm$\,std ($\times100$) over three seeds. MVTec AD~2 uses our matched weak-supervision 80/20 split, not the official normal-only leaderboard protocol. \textbf{Best} is bold and \underline{second best} is underlined.}
\label{tab:combined_comparison}
\end{table*}

%% file: tables/component_analysis.tex
\begin{table}[!t]
\centering
\footnotesize
\setlength{\tabcolsep}{0.2mm}
\begin{tabular*}{0.92\textwidth}{@{\extracolsep{\fill}}lccc@{}}
\toprule
Variant & AUROC-P & AUPRO@.3 & AUPRO@.05 \\
\midrule
GP-free residual pipeline
& $85.4_{\pm2.0}$ & $53.3_{\pm0.8}$ & $27.7_{\pm3.6}$ \\
Spatial GP only ($+G_{\rm GP}$)
& $95.2_{\pm1.3}$ & $79.0_{\pm1.3}$ & $45.5_{\pm1.9}$ \\
Image GP only ($+\widehat p_A$)
& $96.4_{\pm0.6}$ & $82.7_{\pm0.3}$ & $46.1_{\pm1.8}$ \\
\textbf{Full CDGP} (both GP paths)
& $\mathbf{96.8}_{\pm0.7}$ & $\mathbf{84.4}_{\pm0.7}$ & $\mathbf{48.7}_{\pm1.8}$ \\
\bottomrule
\end{tabular*}
\caption{Factorized GP contribution on MVTec AD~2 (mean$\pm$std, $\times100$).
The middle rows each add exactly one GP path to the first row; the first row
contains neither $G_{\rm GP}$ nor $\widehat p_A$.}
\label{tab:component_analysis}
\end{table}

%% file: conclusion.tex
\section{Conclusion}

CDGP combines Dual-GP posterior dominance, hierarchical normal residuals, and training-only calibration for weakly supervised anomaly segmentation. It leads all MVTec AD~2 localization metrics and remains competitive on KSDD2 and VisA. Factorized ablations confirm complementary spatial and image-level roles. Matched linear-head controls delimit the formulation's contribution relative to unconstrained linear scoring. Although the linear kernel makes the training margin linear, joint predictive variance yields a token-dependent dominance score. This supports inducing-variable prediction from image labels without pixel annotations.

%% file: supplement_content.tex
\section{Controlled Evidence on MVTec AD 2}

The controlled analyses follow the primary MVTec AD~2 protocol used for the main-paper ablations. Every experiment uses the same three seed-specific manifests, fixed budgets, and final step-400 evaluation. Metrics are mean$\pm$std ($\times100$).

\subsection{Paired Evidence for the Factorized Contribution}

Main-paper Table~2 separates the Gaussian process (GP)-free residual pipeline, spatial posterior dominance $G_{\rm GP}$, and pooled image evidence $\widehat p_A$. We strengthen that aggregate comparison by pairing the eight category results before inference. Table~\ref{tab:paired_inference} shows that the complete GP path improves all three localization metrics with positive category-bootstrap intervals. Conditional on retaining $\widehat p_A$, the spatial statistic also adds $0.36/1.65/2.65$ points. Posterior dominance is consistently positive relative to the same-model mean margin, while the direct-linear row serves as the stricter parameterization control.

\input{tables/paired_inference_analysis}
\FloatBarrier

\subsection{Posterior Dominance and Functional-Form Controls}

For the linear kernel, CDGP's predictive-mean margin is linear in token space. Table~\ref{tab:linear_control} separates this numerator from the final posterior-dominance statistic. The two CDGP rows use the same trained models, residual maps, splits, and calibration; only the GP statistic entering inference changes. Posterior dominance improves the three localization metrics by $0.1/0.3/0.3$ points. A freely parameterized direct linear head further tests whether the inducing-variable construction adds beyond mean expressivity.

\input{tables/gp_controls_and_fusion}

We also equip a direct linear numerator with an engineered token-dependent quadratic denominator. For each class, the matched control learns
\begin{equation}
\widetilde\sigma_c^2(\mathbf t)
=
\mathbf t^\top\!\operatorname{diag}
\!\left(\operatorname{softplus}(\mathbf a_c)+10^{-4}\right)\mathbf t
+\lVert\mathbf B_c^\top\mathbf t\rVert_2^2+10^{-6},
\label{eq:heteroscedastic_control}
\end{equation}
with $\operatorname{rank}(\mathbf B_N)=32$ and $\operatorname{rank}(\mathbf B_A)=16$, matching the inducing budgets. It uses $(\widetilde\mu_A-\widetilde\mu_N)/ \sqrt{\widetilde\sigma_A^2+\widetilde\sigma_N^2}$ in the same three losses and inference pipeline. Thus the control matches the heteroscedastic linear functional form without GP conditioning, whereas CDGP couples its mean and quadratic normalization through $\mathbf Z_c,\mathbf K_c$, and $\mathbf S_c$.

\input{tables/heteroscedastic_linear_control}

CDGP improves this matched heteroscedastic control by $0.2/0.9/1.0$ points on AUROC-P/AUPRO@.3/AUPRO@.05.

\subsection{Initialization, Inducing-Budget, and Kernel Sensitivity}

We next test the fixed initialization and kernel choices in the complete CDGP pipeline. Table~\ref{tab:initialization_sensitivity} varies the fraction of farthest defective-image tokens retained before farthest point sampling (FPS) and the inducing budgets. Moving the hard-pool fraction from $10\%$ to $40\%$ changes every metric by at most $0.1$ point. Halving $(M_N,M_A)$ to $(16,8)$ produces a small reduction, whereas doubling it to $(64,32)$ provides no average gain and is less stable at low false positive rate (FPR). The default $(32,16)$ is therefore a compact operating point rather than a category-specific choice.

\input{tables/initialization_sensitivity}

Table~\ref{tab:kernel_ablation} compares the default linear kernel with spherical radial basis function (RBF) kernels on the same normalized tokens. The wider RBF settings ($\sigma=.5$ and $1$) closely match the linear kernel, while the narrow $\sigma=.25$ kernel is substantially worse and more variable. Nonlinear kernelization therefore offers no consistent gain in this controlled sweep; we retain the parameter-free linear kernel for its simplicity and stability.

\input{tables/kernel_ablation_final}
\FloatBarrier

\subsection{Inference-Weight Sensitivity}

Table~\ref{tab:fusion_sensitivity} evaluates inference-weight sensitivity without retraining or test-set fitting. The $w=0$ row reports the image-only endpoint; for each positive $w$, the empirical normal tail is re-estimated from its training calibration partition. The default $w=.025$ performs best among the reported local spatial settings, while $\lambda_I=1/3$--$2/3$ forms a broad plateau. These results support a small but nonzero spatial GP weight and show that the image coefficient is not sharply tuned to a single value.

\input{tables/fusion_sensitivity}

\section{Per-Category Results}

Table~\ref{tab:per_category_final} reports the final CDGP pipeline for every category in all three benchmarks. Each entry summarizes three complete runs, and the dataset rows reproduce the aggregation used in the main paper. This disaggregation shows that the reported averages are not driven by one isolated object type.

\input{tables/per_category_results}

\section{Qualitative Results Across All Categories}
\label{sec:qualitative_supp}

Tables~\ref{tab:qualitative_mvtec}--\ref{tab:qualitative_visa} show one defective example from each of the 21 MVTec AD~2, KSDD2, and VisA categories using the final posterior-dominance models.

\input{tables/qualitative_mvtecad2_cells}
\input{tables/qualitative_ksdd2_cells}
\input{tables/qualitative_visa_cells}

\section{Additional Evaluation}

\subsection{Defective-Image Budget Sensitivity}

We retain every normal training image and limit defective images to $k\in\{1,2,4,8,16\}$ per category. A fixed path-seeded permutation makes the subsets nested within every category and seed. Calibration labels are drawn only from the same $k$ labeled defective images; for $k=1$, that image also defines the positive calibration endpoint.

\begin{figure*}[ht!]
  \centering
  \includegraphics[width=.98\textwidth]{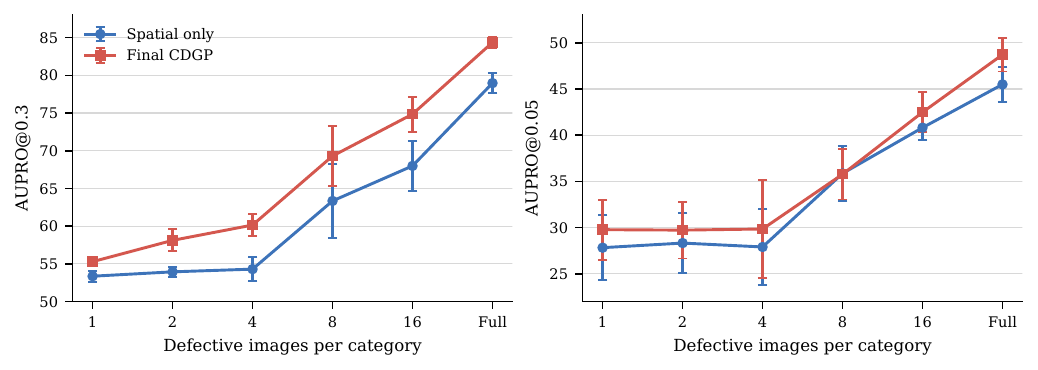}
  \caption{Spatial-only and final-CDGP localization versus defective-image
  budget on MVTec AD~2. Left: AUPRO@.3; right: low-FPR AUPRO@.05. Error bars
  are standard deviations over three seeds.}
  \label{fig:lowshot_budget}
\end{figure*}

\input{tables/lowshot_budget}

Figure~\ref{fig:lowshot_budget} and Table~\ref{tab:lowshot_budget} quantify both spatial and complete-pipeline scaling. Final AUPRO@.3 rises from $55.3$ with one defective image per category to $74.9$ with 16 and $84.4$ with the full weakly labeled split; the corresponding spatial-only values are $53.4$, $68.0$, and $79.0$. Low-FPR AUPRO@.05 exhibits the same overall trend, while explicitly exposing the contribution of broadcast image evidence at each label budget. AUROC-I rises from $69.6$ with one defective image to $95.6$ with 16 and $100.0$ with the full split.

\subsection{Inference Complexity}

Table~\ref{tab:complexity} reports the complete batch-one pipeline at $256\!\times\!256$. CDGP runs in $7.07$ ms per image (approximately 141 images/s) with 357.4 MiB peak memory on an RTX 6000 Ada.

\input{tables/complexity}

\section{Reproducibility Details}

\subsection{Final Configuration and Data Provenance}

Table~\ref{tab:reproduction_hyperparameters} lists all settings required to reproduce the final method. Table~\ref{tab:split_provenance} gives the exact split construction. Seed/category manifests store resolved image paths and SHA-256 digests, and an automated audit verifies zero calibration--test overlap. Training datasets are rebuilt with mask paths set to \texttt{None}; the residual student receives only normal fit images, whereas the logistic calibrator receives only binary labels from the disjoint calibration subset.

\input{tables/reproducibility_details}

\paragraph{Randomness and repetitions.}
Python, NumPy, PyTorch, and all CUDA devices are seeded with $s\in\{0,1,2\}$. Metrics are first averaged over the eight categories within a seed and then summarized across seeds. The label-budget selector derives one category/seed-specific permutation from a SHA-256 path seed and takes nested prefixes of length $1,2,4,8,$ and $16$; its manifest records the selected paths, effective budget, calibration subset, and both manifest hashes.

\paragraph{Metric implementation.}
AUROC-I uses image labels and $\widehat p_A$; AUROC-P pools all pixels. AUPRO uses 8-connected ground-truth components, sorts all pixel scores (no threshold subsampling), linearly interpolates the PRO curve at the specified maximum FPR, and divides the trapezoidal integral by that maximum. We report $\mathrm{FPR}_{\max}=.3$ and $.05$; the latter emphasizes the industrial low-false-alarm regime.

\paragraph{Hardware and software.}
Experiments ran on NVIDIA RTX 6000 Ada GPUs (48~GiB, driver 570.207), dual AMD EPYC 9124 CPUs (32 physical cores), and 1.5~TiB RAM under Ubuntu 24.04 and Linux 6.8. The environment uses Python 3.12.3, PyTorch 2.7.0, torchvision 0.22.0, CUDA 12.8, cuDNN 9.8, timm 1.0.26, NumPy 1.26.4, SciPy 1.11.4, scikit-learn 1.4.1.post1, pandas 2.1.4, and Pillow 10.2.0. Training/evaluation scripts, preprocessing code, manifests, exact metric code, configs, and analysis scripts will be released upon publication.

%% file: tables/paired_inference_analysis.tex
\par\vspace{1.0\baselineskip}
\noindent\begin{minipage}{\columnwidth}
\centering
\footnotesize
\setlength{\tabcolsep}{1.0mm}
\renewcommand{\arraystretch}{0.92}
\begin{tabular*}{0.92\textwidth}{@{\extracolsep{\fill}}ll@{}}
\toprule
Paired contrast / metric & Difference $[95\%\ \mathrm{CI}]$ ($p$) \\
\midrule
Full CDGP $-$ GP-free / AUROC-P & $+11.38\ [7.22,16.71]\ (.023)$ \\
\quad / AUPRO@.3 & $+31.02\ [22.23,40.32]\ (.023)$ \\
\quad / AUPRO@.05 & $+20.97\ [14.98,25.39]\ (.023)$ \\
\addlinespace[1pt]
Spatial $G_{\rm GP}$ (image fixed) / AUROC-P & $+0.36\ [0.17,0.58]\ (.023)$ \\
\quad / AUPRO@.3 & $+1.65\ [0.88,2.50]\ (.023)$ \\
\quad / AUPRO@.05 & $+2.65\ [1.55,3.73]\ (.023)$ \\
\addlinespace[1pt]
Posterior $-$ mean margin / AUROC-P & $+0.05\ [0.02,0.10]\ (.031)$ \\
\quad / AUPRO@.3 & $+0.28\ [0.12,0.49]\ (.023)$ \\
\quad / AUPRO@.05 & $+0.30\ [0.12,0.51]\ (.031)$ \\
\addlinespace[1pt]
Posterior $-$ direct linear / AUROC-P & $+0.16\ [0.03,0.28]\ (.141)$ \\
\quad / AUPRO@.3 & $+0.52\ [0.18,0.86]\ (.141)$ \\
\quad / AUPRO@.05 & $+0.59\ [-0.01,1.13]\ (.141)$ \\
\bottomrule
\end{tabular*}
\captionof{table}{Paired MVTec AD~2 category analysis (percentage points),
reported as difference [95\% category-bootstrap interval] after seed
averaging. Intervals use 100,000 resamples; $p$ uses an exact two-sided
paired sign-flip test and is Holm-adjusted within each three-metric
contrast.}
\label{tab:paired_inference}
\end{minipage}

%% file: tables/gp_controls_and_fusion.tex
\begin{table}[ht!]
\centering
\footnotesize
\setlength{\tabcolsep}{0.5mm}
\begin{tabular*}{0.92\textwidth}{@{\extracolsep{\fill}}lccc@{}}
\toprule
Variant & AUROC-P & AUPRO@.3 & AUPRO@.05 \\
\midrule
Direct linear head
& $96.6_{\pm0.7}$ & $83.8_{\pm0.4}$ & $48.0_{\pm1.9}$ \\
CDGP mean margin $s_\mu$
& $96.7_{\pm0.7}$ & $84.1_{\pm0.7}$ & $48.4_{\pm2.0}$ \\
CDGP posterior dominance $G_{\rm GP}$
& $96.8_{\pm0.7}$ & $84.4_{\pm0.7}$ & $48.7_{\pm1.8}$ \\
\bottomrule
\end{tabular*}
\caption{Matched scoring and parameterization controls on MVTec AD~2
(mean$\pm$std, $\times100$). The two CDGP rows use the same trained models and differ
only in inference scoring.}
\label{tab:linear_control}
\end{table}

%% file: tables/heteroscedastic_linear_control.tex
\begin{table}[ht!]
\centering
\footnotesize
\setlength{\tabcolsep}{0.55mm}
\begin{tabular*}{0.92\textwidth}{@{\extracolsep{\fill}}lccc@{}}
\toprule
Variant & AUROC-P & AUPRO@.3 & AUPRO@.05 \\
\midrule
Heteroscedastic linear
& $96.6_{\pm0.7}$ & $83.5_{\pm0.5}$ & $47.7_{\pm2.3}$ \\
CDGP
& $96.8_{\pm0.7}$ & $84.4_{\pm0.7}$ & $48.7_{\pm1.8}$ \\
\bottomrule
\end{tabular*}
\caption{Strong functional-form control. The direct model learns two free linear means and two diagonal-plus-low-rank positive-semidefinite quadratic
variances.}
\label{tab:heteroscedastic_linear_control}
\end{table}

%% file: tables/initialization_sensitivity.tex
\begin{table}[ht!]
\centering
\footnotesize
\setlength{\tabcolsep}{1.2mm}
\begin{tabular*}{0.92\textwidth}{@{\extracolsep{\fill}}llccc@{}}
\toprule
Factor & Setting & AUROC-P & AUPRO@.3 & AUPRO@.05 \\
\midrule
Hard-pool fraction & 10\% & $96.8_{\pm0.7}$ & $84.5_{\pm0.6}$ & $48.8_{\pm1.6}$ \\
 & 20\% (default) & $96.8_{\pm0.7}$ & $84.4_{\pm0.7}$ & $48.7_{\pm1.8}$ \\
 & 40\% & $96.8_{\pm0.7}$ & $84.4_{\pm0.6}$ & $48.7_{\pm1.8}$ \\
\midrule
Inducing budgets $(M_N,M_A)$ & $(16,8)$ & $96.6_{\pm0.8}$ & $83.9_{\pm0.6}$ & $48.3_{\pm2.2}$ \\
 & $(32,16)$ (default) & $96.8_{\pm0.7}$ & $84.4_{\pm0.7}$ & $48.7_{\pm1.8}$ \\
 & $(64,32)$ & $96.7_{\pm0.5}$ & $84.3_{\pm0.3}$ & $47.5_{\pm2.2}$ \\
\bottomrule
\end{tabular*}
\caption{One-factor sensitivity of the CDGP pipeline on MVTec AD~2. Hard-pool fraction changes only the farthest-token candidate percentile; inducing-budget rows change only $(M_N,M_A)$.}
\label{tab:initialization_sensitivity}
\end{table}

%% file: tables/kernel_ablation_final.tex
\begin{table}[ht!]
\centering
\footnotesize
\setlength{\tabcolsep}{1.4mm}
\begin{tabular*}{0.92\textwidth}{@{\extracolsep{\fill}}lccc@{}}
\toprule
Setting & AUROC-P & AUPRO@.3 & AUPRO@.05 \\
\midrule
Linear (default) & $96.8_{\pm0.7}$ & $84.4_{\pm0.7}$ & $48.7_{\pm1.8}$ \\
Spherical RBF, $\sigma=.25$ & $94.7_{\pm2.2}$ & $76.6_{\pm7.3}$ & $42.8_{\pm2.3}$ \\
Spherical RBF, $\sigma=.5$ & $96.7_{\pm0.7}$ & $84.1_{\pm0.5}$ & $48.3_{\pm1.9}$ \\
Spherical RBF, $\sigma=1$ & $96.7_{\pm0.8}$ & $84.4_{\pm0.5}$ & $48.5_{\pm2.1}$ \\
\bottomrule
\end{tabular*}
\caption{Kernel control for the CDGP pipeline on MVTec AD~2. RBF rows use $k(\mathbf z,\mathbf z')=\exp[-(1-\mathbf z^\top\mathbf z')/\sigma^2]$ on normalized tokens.}
\label{tab:kernel_ablation}
\end{table}

%% file: tables/fusion_sensitivity.tex
\begin{table}[ht!]
\centering
\footnotesize
\setlength{\tabcolsep}{1.0mm}

\begin{tabular*}{0.92\textwidth}{@{\extracolsep{\fill}}cccc@{}}
\toprule
GP weight $w$ & AUROC-P & AUPRO@.3 & AUPRO@.05 \\
\midrule
$0$ & $96.42_{\pm0.62}$ & $82.72_{\pm0.31}$ & $46.07_{\pm1.76}$ \\
$0.01$  & $96.65_{\pm0.66}$ & $83.64_{\pm0.41}$ & $47.93_{\pm1.87}$ \\
$0.025$ (default) & $96.78_{\pm0.70}$ & $84.36_{\pm0.74}$ & $48.69_{\pm1.80}$ \\
$0.05$  & $96.22_{\pm0.64}$ & $83.82_{\pm1.53}$ & $45.28_{\pm1.10}$ \\
\bottomrule
\end{tabular*}
\par\vspace{0.8mm}
\textbf{(a) Spatial fusion weight}\par\vspace{1mm}
\par\vspace{2mm}

\begin{tabular*}{0.92\textwidth}{@{\extracolsep{\fill}}cccc@{}}
\toprule
Image weight $\lambda_I$ & AUROC-P & AUPRO@.3 & AUPRO@.05 \\
\midrule
$0$ & $96.40_{\pm0.60}$ & $82.70_{\pm0.30}$ & $46.10_{\pm1.80}$ \\
$1/6$ & $96.58_{\pm0.75}$ & $83.52_{\pm0.78}$ & $48.15_{\pm1.86}$ \\
$1/3$ (default) & $96.78_{\pm0.70}$ & $84.36_{\pm0.73}$ & $48.71_{\pm1.83}$ \\
$1/2$ & $96.81_{\pm0.70}$ & $84.52_{\pm0.73}$ & $48.82_{\pm1.81}$ \\
$2/3$ & $96.82_{\pm0.69}$ & $84.55_{\pm0.73}$ & $48.86_{\pm1.81}$ \\
\bottomrule
\end{tabular*}
\par\vspace{0.8mm}
\textbf{(b) Broadcast image weight}
\caption{Inference-weight sensitivity on MVTec AD~2. (a) varies the
GP--residual weight. The $w=0$ row reports the image-only endpoint;
positive-$w$ rows use separately fitted training-only normal tails. Every
row retains $\widehat p_A$, so $w=0$ is not the GP-free pipeline.
(b) varies the broadcast image coefficient at fixed $w=.025$.}
\label{tab:fusion_sensitivity}
\label{tab:image_weight_sensitivity}
\end{table}

%% file: tables/per_category_results.tex
\begin{table*}[ht!]
\centering
\footnotesize
\setlength{\tabcolsep}{2.0mm}
\begin{tabular*}{0.92\textwidth}{@{\extracolsep{\fill}}llcccc@{}}
\toprule
Dataset & Category & AUROC-I & AUROC-P & AUPRO@.3 & AUPRO@.05 \\
\midrule
MVTec AD~2 & can & $100.0_{\pm0.0}$ & $97.5_{\pm0.3}$ & $89.1_{\pm0.5}$ & $42.1_{\pm3.4}$ \\
 & fabric & $100.0_{\pm0.0}$ & $95.1_{\pm0.3}$ & $77.9_{\pm3.5}$ & $25.3_{\pm3.6}$ \\
 & fruit\_jelly & $100.0_{\pm0.0}$ & $98.5_{\pm0.4}$ & $94.2_{\pm1.5}$ & $67.3_{\pm9.8}$ \\
 & rice & $100.0_{\pm0.0}$ & $96.0_{\pm0.9}$ & $87.2_{\pm2.5}$ & $53.8_{\pm5.5}$ \\
 & sheet\_metal & $100.0_{\pm0.0}$ & $92.5_{\pm3.9}$ & $48.1_{\pm6.0}$ & $10.5_{\pm1.3}$ \\
 & vial & $100.0_{\pm0.0}$ & $98.4_{\pm0.1}$ & $97.3_{\pm0.3}$ & $83.6_{\pm1.8}$ \\
 & wall\_plugs & $100.0_{\pm0.0}$ & $97.4_{\pm0.6}$ & $88.2_{\pm0.7}$ & $43.5_{\pm3.6}$ \\
 & walnuts & $100.0_{\pm0.0}$ & $98.8_{\pm0.9}$ & $93.1_{\pm1.1}$ & $63.6_{\pm4.2}$ \\
 & \textbf{Mean} & $\mathbf{100.0}_{\pm0.0}$ & $\mathbf{96.8}_{\pm0.7}$ & $\mathbf{84.4}_{\pm0.7}$ & $\mathbf{48.7}_{\pm1.8}$ \\
\midrule
KSDD2 & KSDD2 & $97.4_{\pm0.2}$ & $98.0_{\pm0.6}$ & $93.5_{\pm0.1}$ & -- \\
 & \textbf{Mean} & $\mathbf{97.4}_{\pm0.2}$ & $\mathbf{98.0}_{\pm0.6}$ & $\mathbf{93.5}_{\pm0.1}$ & -- \\
\midrule
VisA & candle & $99.9_{\pm0.0}$ & $99.5_{\pm0.3}$ & $99.3_{\pm0.0}$ & -- \\
 & capsules & $93.6_{\pm0.6}$ & $99.2_{\pm0.1}$ & $88.7_{\pm1.4}$ & -- \\
 & cashew & $97.3_{\pm0.1}$ & $99.4_{\pm0.0}$ & $97.9_{\pm0.1}$ & -- \\
 & chewinggum & $99.6_{\pm0.4}$ & $99.8_{\pm0.0}$ & $94.6_{\pm0.6}$ & -- \\
 & fryum & $100.0_{\pm0.0}$ & $98.7_{\pm0.2}$ & $97.7_{\pm0.3}$ & -- \\
 & macaroni1 & $99.5_{\pm0.0}$ & $99.9_{\pm0.0}$ & $98.9_{\pm0.3}$ & -- \\
 & macaroni2 & $98.6_{\pm0.6}$ & $99.7_{\pm0.1}$ & $97.5_{\pm0.6}$ & -- \\
 & pcb1 & $99.5_{\pm0.0}$ & $99.6_{\pm0.0}$ & $96.8_{\pm0.1}$ & -- \\
 & pcb2 & $98.6_{\pm0.9}$ & $99.0_{\pm0.2}$ & $96.2_{\pm1.3}$ & -- \\
 & pcb3 & $95.2_{\pm3.5}$ & $98.9_{\pm1.2}$ & $92.4_{\pm8.6}$ & -- \\
 & pcb4 & $99.9_{\pm0.0}$ & $99.0_{\pm0.0}$ & $95.7_{\pm0.2}$ & -- \\
 & pipe\_fryum & $99.5_{\pm0.4}$ & $99.6_{\pm0.0}$ & $98.1_{\pm0.3}$ & -- \\
 & \textbf{Mean} & $\mathbf{98.4}_{\pm0.3}$ & $\mathbf{99.4}_{\pm0.1}$ & $\mathbf{96.1}_{\pm0.7}$ & -- \\
\bottomrule
\end{tabular*}
\caption{CDGP results by dataset category. Entries are mean$\pm$std over three complete runs ($\times100$). Dataset means are computed within each seed before the three-seed summary, matching the main-paper aggregation. AUPRO@.05 is reported for the primary MVTec AD~2 benchmark.}
\label{tab:per_category_final}
\end{table*}

%% file: tables/qualitative_mvtecad2_cells.tex
\begin{table*}[ht!]
\centering
\scriptsize
\setlength{\tabcolsep}{0.55mm}
\renewcommand{\arraystretch}{0.82}
\newcommand{\qcellm}[1]{\includegraphics[width=.162\textwidth,height=.108\textwidth,keepaspectratio]{figure/supplement_qualitative_cells_all/mvtecad2_#1}}
\begin{tabular}{@{}lccccc@{}}
\toprule
Category & Raw & Ground truth & Normal mean & Anomaly mean & Final CDGP \\
\midrule
Can & \qcellm{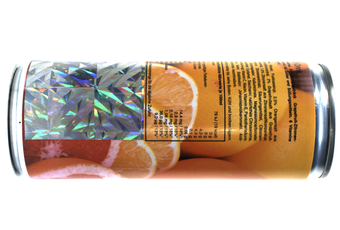} & \qcellm{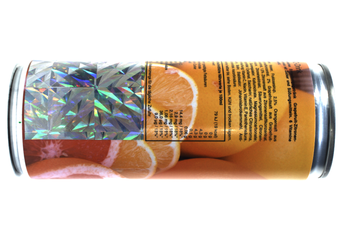} & \qcellm{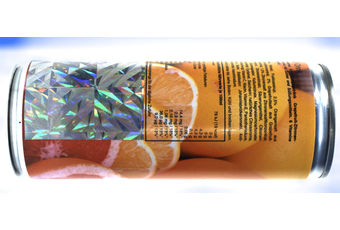} & \qcellm{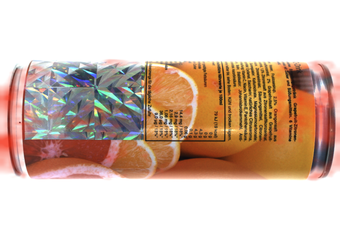} & \qcellm{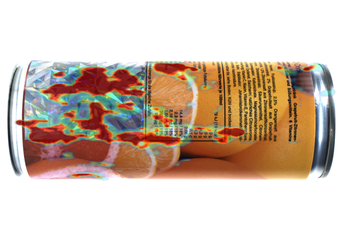} \\
Fabric & \qcellm{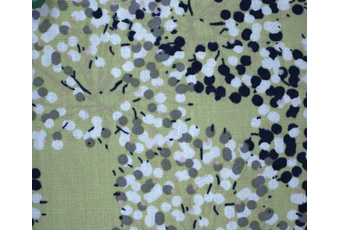} & \qcellm{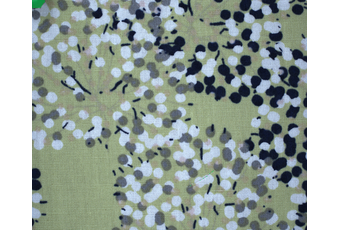} & \qcellm{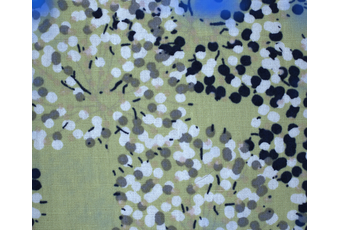} & \qcellm{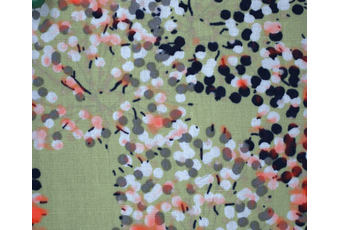} & \qcellm{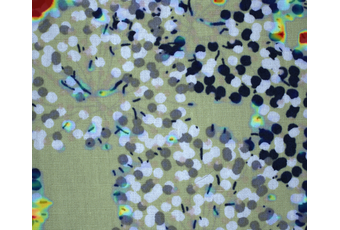} \\
Fruit jelly & \qcellm{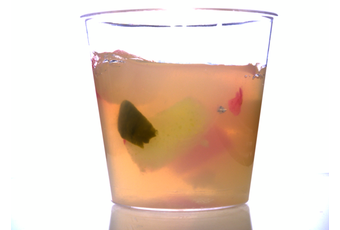} & \qcellm{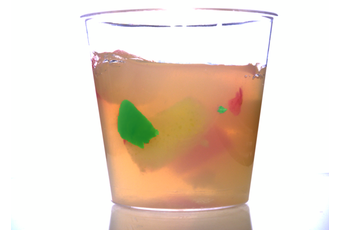} & \qcellm{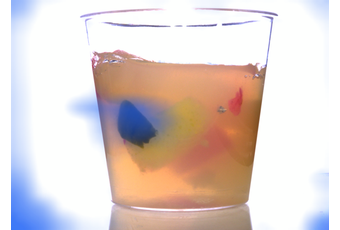} & \qcellm{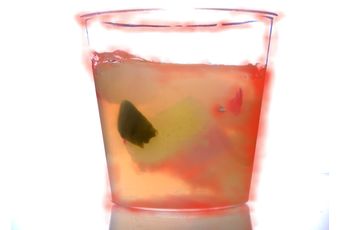} & \qcellm{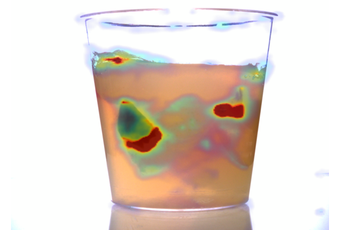} \\
Rice & \qcellm{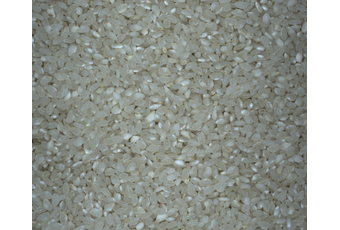} & \qcellm{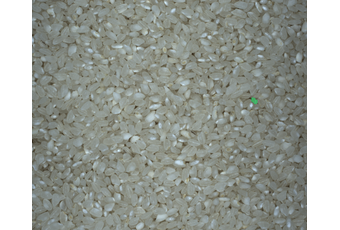} & \qcellm{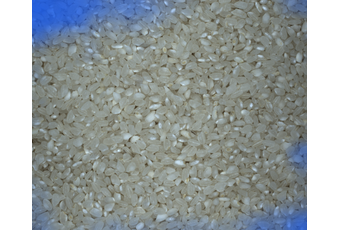} & \qcellm{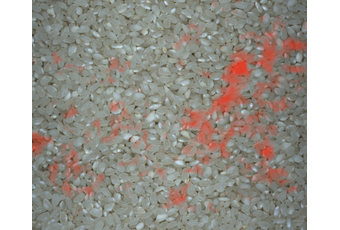} & \qcellm{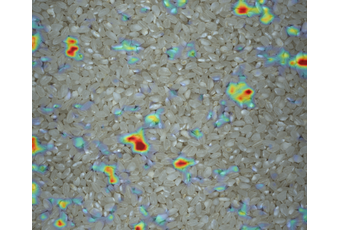} \\
Sheet metal & \qcellm{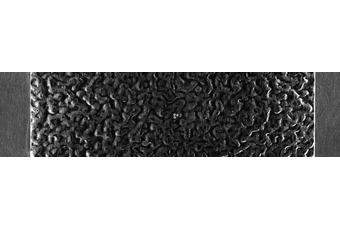} & \qcellm{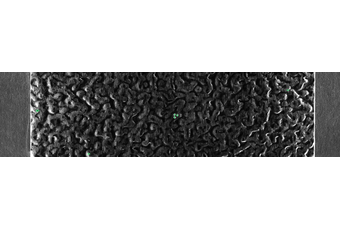} & \qcellm{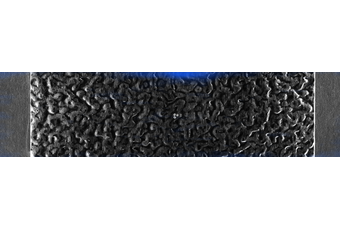} & \qcellm{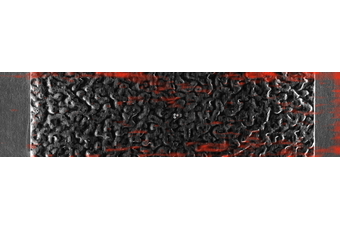} & \qcellm{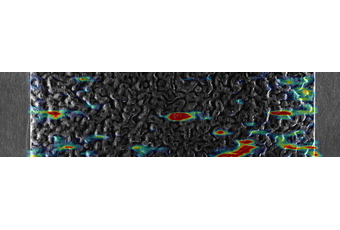} \\
Vial & \qcellm{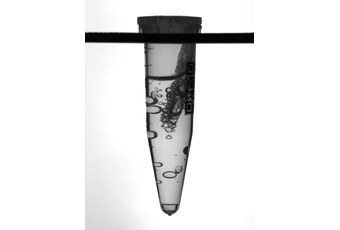} & \qcellm{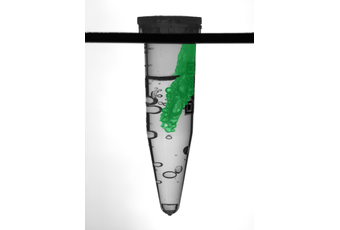} & \qcellm{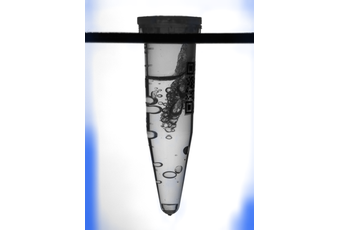} & \qcellm{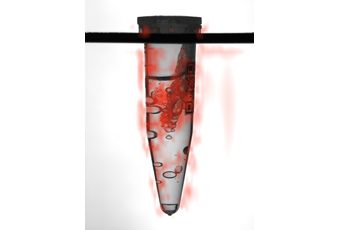} & \qcellm{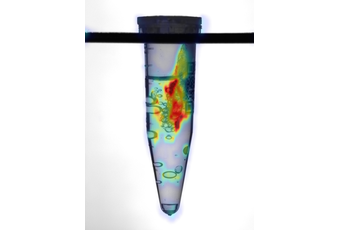} \\
Wall plugs & \qcellm{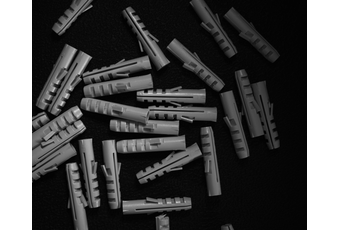} & \qcellm{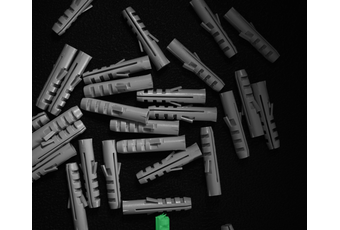} & \qcellm{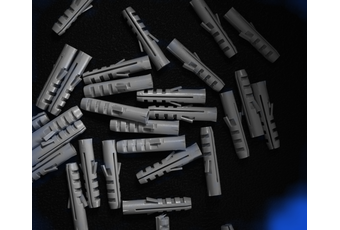} & \qcellm{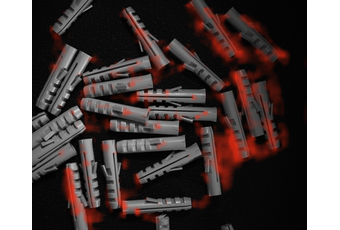} & \qcellm{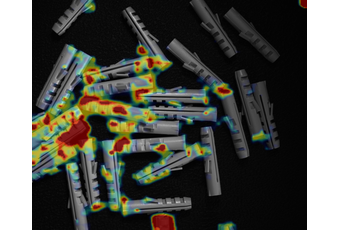} \\
Walnuts & \qcellm{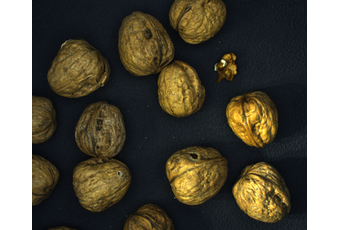} & \qcellm{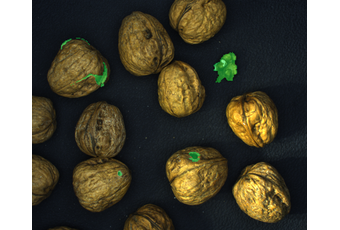} & \qcellm{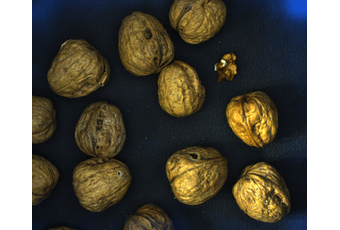} & \qcellm{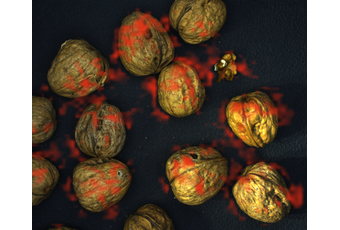} & \qcellm{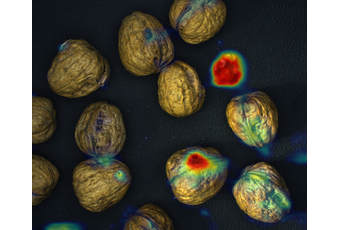} \\
\bottomrule
\end{tabular}
\caption{MVTec AD~2 qualitative decomposition, one defective example per
category.}
\label{tab:qualitative_mvtec}
\end{table*}

%% file: tables/qualitative_ksdd2_cells.tex
\begin{table*}[ht!]
\centering
\scriptsize
\setlength{\tabcolsep}{0.55mm}
\newcommand{\qcellk}[1]{\includegraphics[width=.162\textwidth,height=.145\textwidth,keepaspectratio]{figure/supplement_qualitative_cells_all/ksdd2_ksdd2_#1}}
\begin{tabular}{@{}lccccc@{}}
\toprule
Dataset & Raw & Ground truth & Normal mean & Anomaly mean & Final CDGP \\
\midrule
KSDD2 & \qcellk{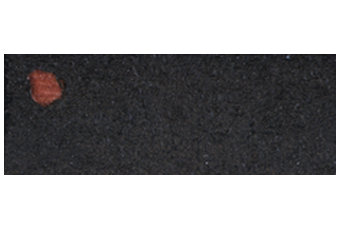} & \qcellk{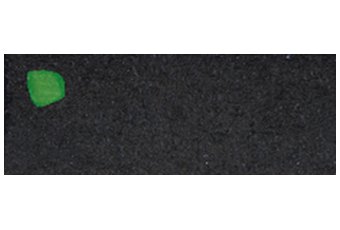} & \qcellk{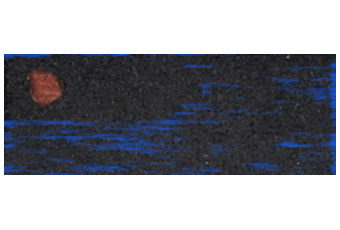} & \qcellk{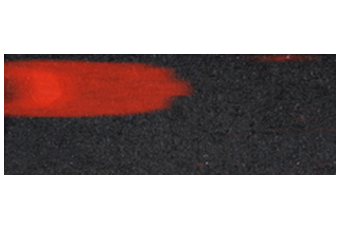} & \qcellk{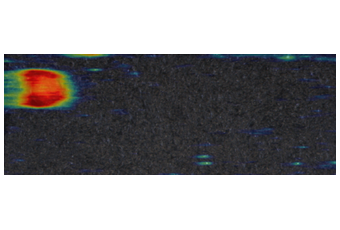} \\
\bottomrule
\end{tabular}
\caption{KSDD2 qualitative decomposition.}
\label{tab:qualitative_ksdd2}
\end{table*}

%% file: tables/qualitative_visa_cells.tex
\begin{table*}[ht!]
\centering
\scriptsize
\setlength{\tabcolsep}{0.55mm}
\renewcommand{\arraystretch}{0.78}
\newcommand{\qcellv}[1]{\includegraphics[width=.162\textwidth,height=.071\textwidth,keepaspectratio]{figure/supplement_qualitative_cells_all/visa_#1}}
\begin{tabular}{@{}lccccc@{}}
\toprule
Category & Raw & Ground truth & Normal mean & Anomaly mean & Final CDGP \\
\midrule
Candle & \qcellv{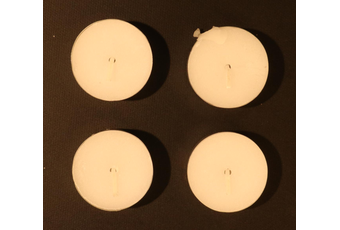} & \qcellv{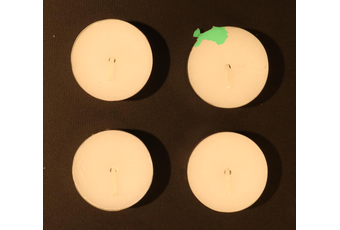} & \qcellv{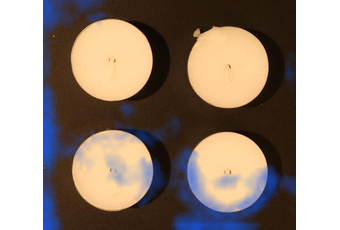} & \qcellv{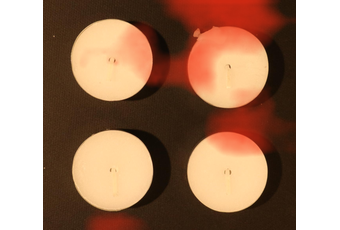} & \qcellv{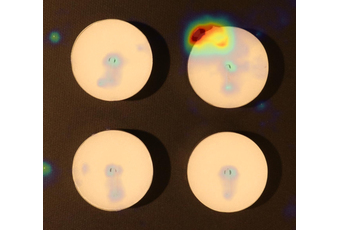} \\
Capsules & \qcellv{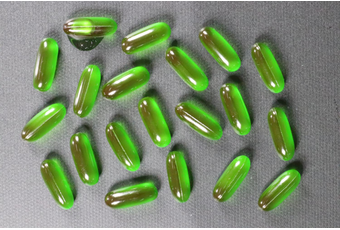} & \qcellv{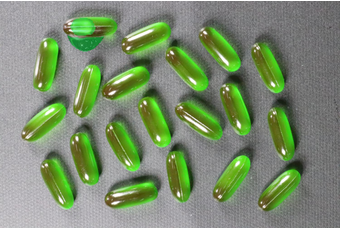} & \qcellv{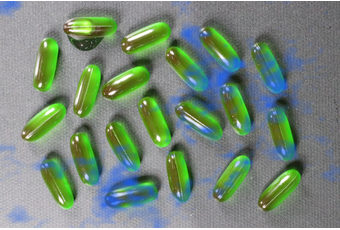} & \qcellv{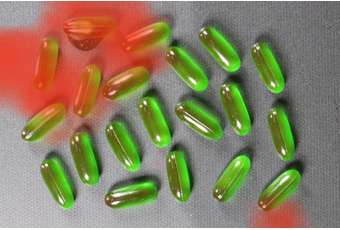} & \qcellv{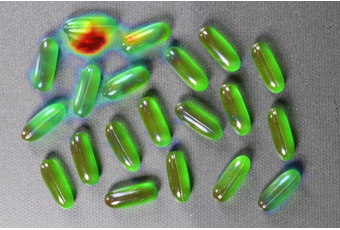} \\
Cashew & \qcellv{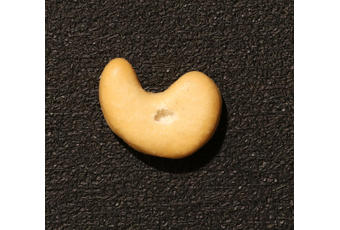} & \qcellv{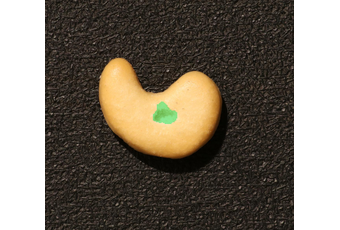} & \qcellv{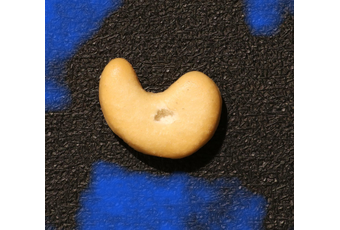} & \qcellv{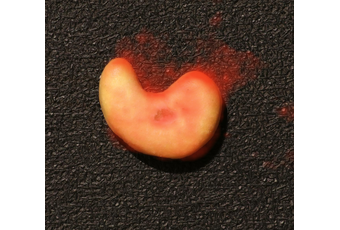} & \qcellv{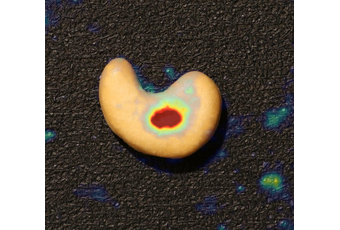} \\
Chewing gum & \qcellv{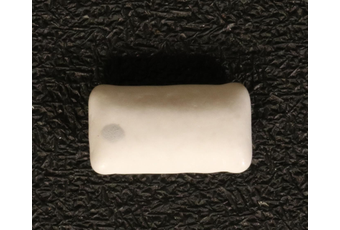} & \qcellv{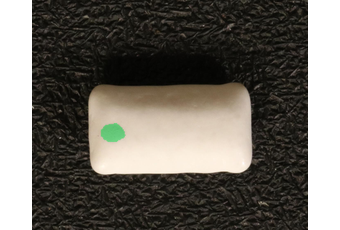} & \qcellv{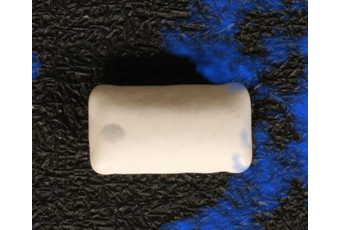} & \qcellv{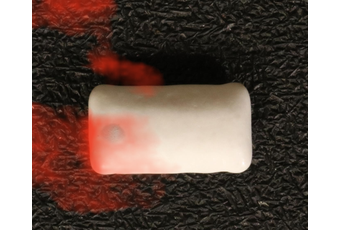} & \qcellv{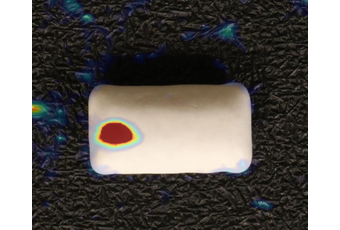} \\
Fryum & \qcellv{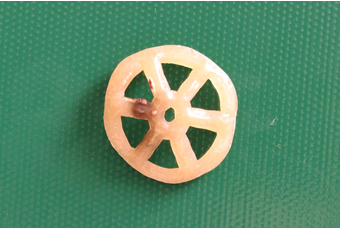} & \qcellv{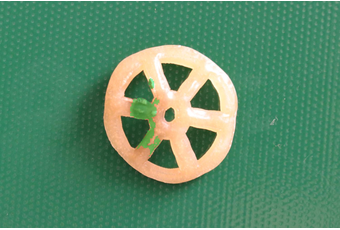} & \qcellv{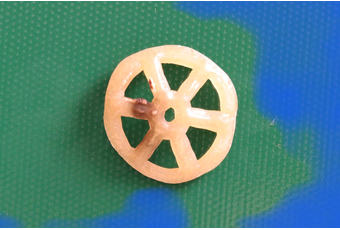} & \qcellv{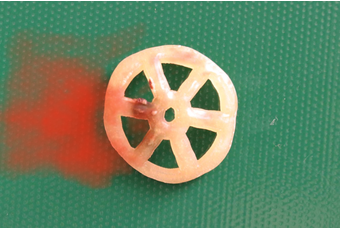} & \qcellv{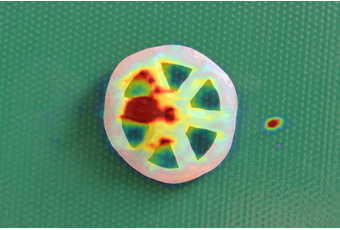} \\
Macaroni 1 & \qcellv{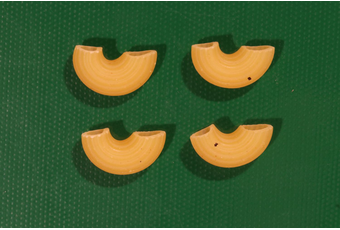} & \qcellv{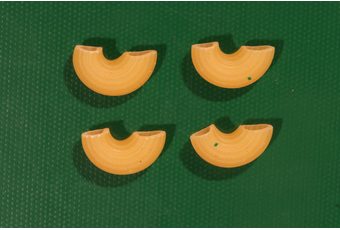} & \qcellv{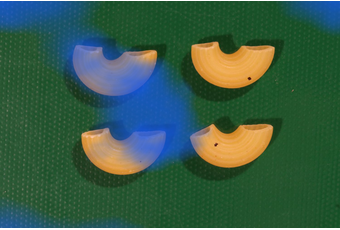} & \qcellv{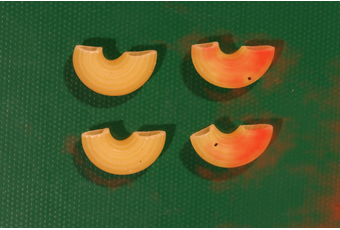} & \qcellv{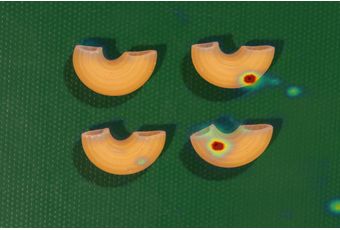} \\
Macaroni 2 & \qcellv{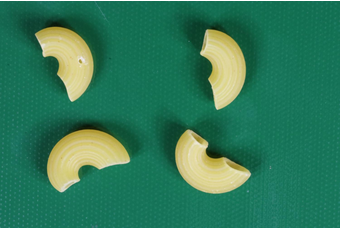} & \qcellv{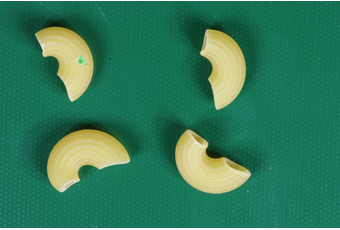} & \qcellv{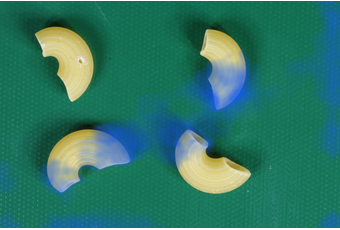} & \qcellv{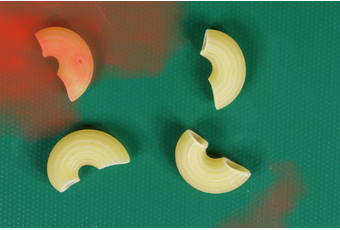} & \qcellv{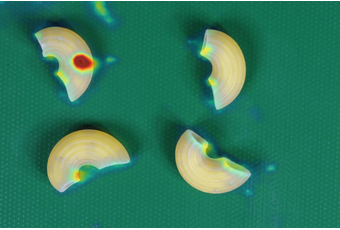} \\
PCB 1 & \qcellv{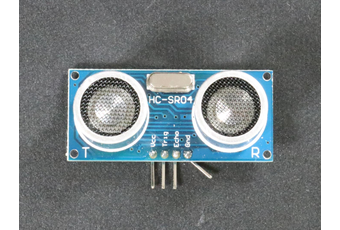} & \qcellv{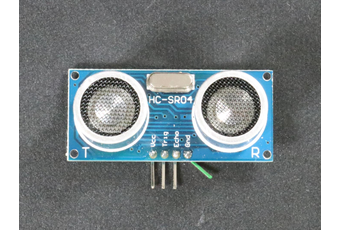} & \qcellv{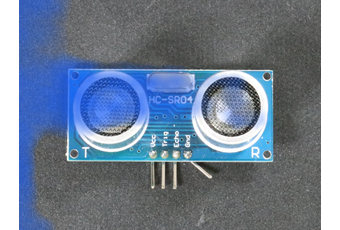} & \qcellv{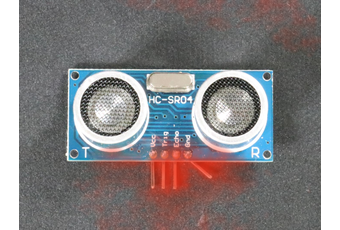} & \qcellv{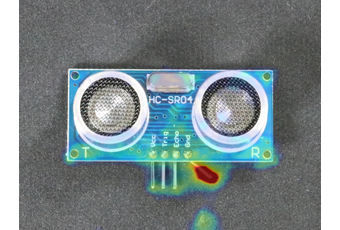} \\
PCB 2 & \qcellv{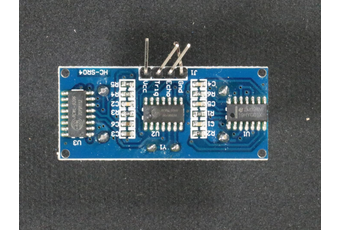} & \qcellv{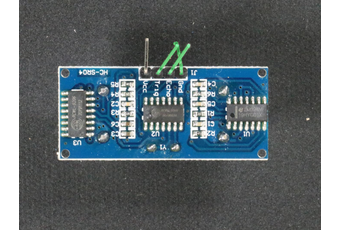} & \qcellv{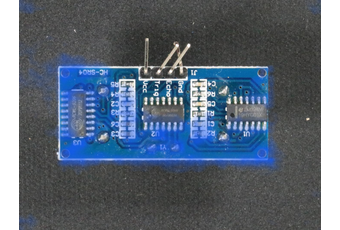} & \qcellv{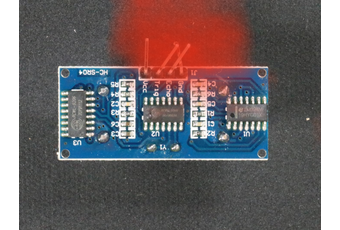} & \qcellv{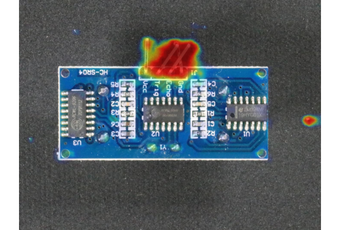} \\
PCB 3 & \qcellv{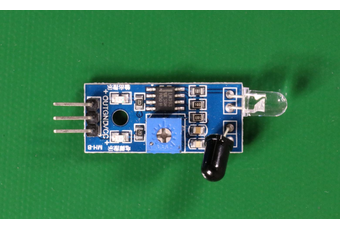} & \qcellv{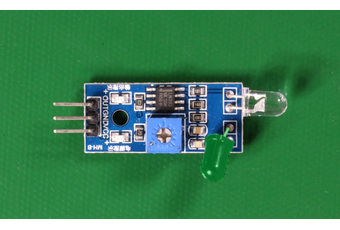} & \qcellv{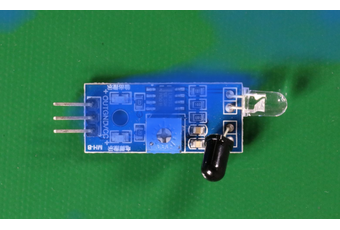} & \qcellv{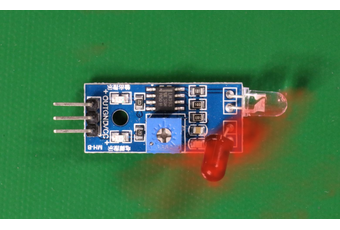} & \qcellv{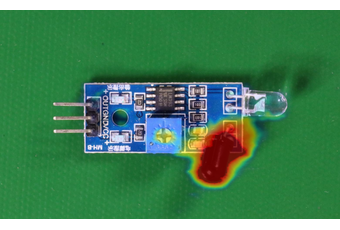} \\
PCB 4 & \qcellv{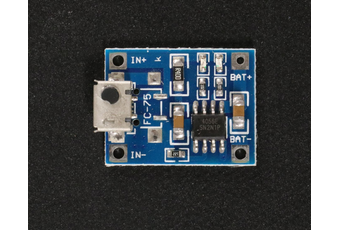} & \qcellv{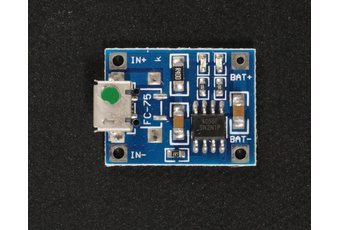} & \qcellv{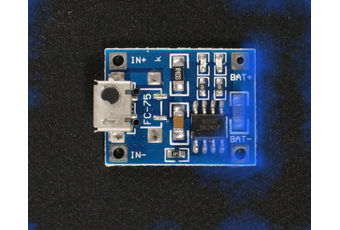} & \qcellv{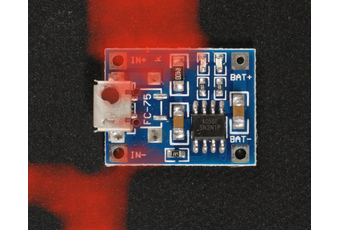} & \qcellv{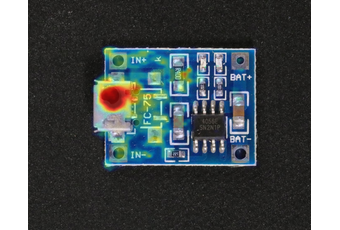} \\
Pipe fryum & \qcellv{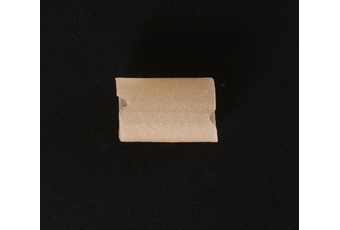} & \qcellv{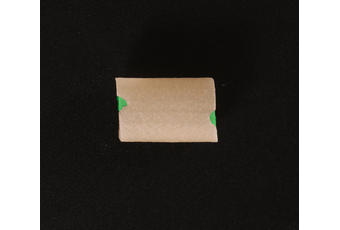} & \qcellv{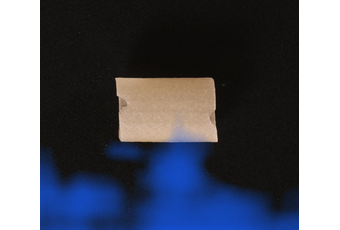} & \qcellv{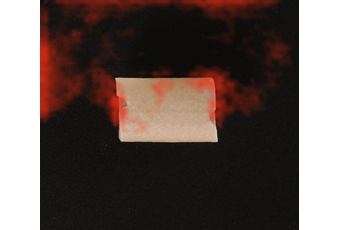} & \qcellv{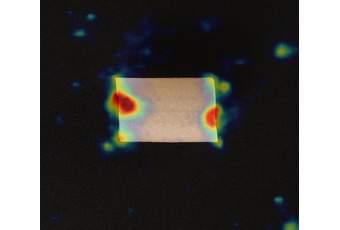} \\
\bottomrule
\end{tabular}
\caption{VisA qualitative decomposition, one defective example per category.}
\label{tab:qualitative_visa}
\end{table*}

%% file: tables/lowshot_budget.tex
\begin{table*}[ht!]
\centering
\footnotesize
\setlength{\tabcolsep}{3.5mm}
\renewcommand{\arraystretch}{1.08}
\begin{tabular*}{0.92\textwidth}{@{\extracolsep{\fill}}lccccc@{}}
\toprule
& & \multicolumn{2}{c}{AUPRO@.3} & \multicolumn{2}{c}{AUPRO@.05} \\
\cmidrule(lr){3-4}\cmidrule(lr){5-6}
Defect images/category & AUROC-I & Spatial & Final & Spatial & Final \\
\midrule
1 & $69.6_{\pm4.3}$ & $53.4_{\pm0.7}$ & $55.3_{\pm0.3}$ & $27.8_{\pm3.5}$ & $29.8_{\pm3.2}$ \\
2 & $76.6_{\pm5.8}$ & $53.9_{\pm0.7}$ & $58.1_{\pm1.5}$ & $28.3_{\pm3.2}$ & $29.7_{\pm3.1}$ \\
4 & $85.2_{\pm3.4}$ & $54.3_{\pm1.6}$ & $60.2_{\pm1.4}$ & $27.9_{\pm4.1}$ & $29.9_{\pm5.3}$ \\
8 & $93.2_{\pm2.5}$ & $63.3_{\pm4.9}$ & $69.3_{\pm4.0}$ & $35.9_{\pm3.0}$ & $35.8_{\pm2.7}$ \\
16 & $95.6_{\pm0.2}$ & $68.0_{\pm3.3}$ & $74.9_{\pm2.3}$ & $40.8_{\pm1.3}$ & $42.5_{\pm2.2}$ \\
Full & $100.0_{\pm0.0}$ & $79.0_{\pm1.3}$ & $84.4_{\pm0.7}$ & $45.5_{\pm1.9}$ & $48.7_{\pm1.8}$ \\
\bottomrule
\end{tabular*}
\caption{Defective-image budget sensitivity on MVTec AD~2. AUROC-I
evaluates image anomaly probability; Spatial retains the calibrated spatial
map but omits broadcast image evidence; Final is complete CDGP.}
\label{tab:lowshot_budget}
\end{table*}

%% file: tables/complexity.tex
\begin{table*}[ht!]
\centering
\footnotesize
\setlength{\tabcolsep}{1.4mm}

\begin{tabular*}{0.92\textwidth}{@{\extracolsep{\fill}}lccc@{}}
\toprule
Model & Params. (M) & Latency (ms) & Peak VRAM (MiB) \\
\midrule
CDGP      & 63.8 & $7.07\!\pm\!0.32$ & 357.4 \\
\bottomrule
\end{tabular*}
\caption{Inference complexity at $256\!\times\!256$. Latency is measured with batch size one on an RTX 6000 Ada after 10 warm-up iterations and over 50 timed iterations.}
\label{tab:complexity}
\end{table*}

%% file: tables/reproducibility_details.tex
\begin{table*}[ht!]
\centering
\footnotesize
\setlength{\tabcolsep}{2.0mm}
\begin{tabular}{@{}p{0.16\textwidth}p{0.77\textwidth}@{}}
\toprule
Component & Final setting \\
\midrule
Input and encoder
& RGB bilinear resize to $256\!\times\!256$ (MVTec AD~2/VisA) or
$256\!\times\!704$ (KSDD2); ImageNet mean/std normalization; ImageNet-1K V2
WideResNet-50-2 weights; layer1--3; $D=256$; training-only color jitter
$(.12,.12,.05,.02)$. \\
Dual GP
& Linear kernel; $M_N=32$, $M_A=16$; hard pool $20\%$; jitter $10^{-3}$;
variance floor $10^{-6}$; batch 16; 400 steps; AdamW, base LR $10^{-4}$,
inducing-value LR $10^{-3}$, weight decay $10^{-4}$ except zero on inducing
values; cosine decay to $0.05$ of the initial LR; margin $\gamma=.5$;
$(\mathcal L_{\rm MIL},\mathcal L_{\rm CMP},\mathcal L_{\rm ABN})$ weights
$(1,1,4)$. \\
Residual student
& Normal fit images only; batch 5 (MVTec AD~2/VisA) or 3 (KSDD2); 360
steps; AdamW LR $3\!\times\!10^{-4}$, weight decay $10^{-4}$, cosine decay
to $10^{-6}$; three scales; hard fraction $q=.10$,
$(\lambda_h,\lambda_s)=(.65,.10)$. \\
Normal corruption
& Gaussian noise std $.035$; $3\!\times\!3$ local averaging with probability
$.70$; zero rectangle with probability $.75$, with each side sampled between
approximately $1/16$ and $1/5$ of the corresponding image dimension. \\
Calibration/\allowbreak inference
& Training fit/calibration split $80/20$; $w=.025$; identity spatial
operator (no Gaussian smoothing or max blend); at most $10^6$ normal
calibration pixels sampled deterministically; tail floor $e^{-16}$; LSE
$r=8$; logistic $C=1$, balanced classes, nonnegative slope;
$\lambda_I=1/3$. \\
Evaluation
& Seeds $\{0,1,2\}$; final model at step 400; category mean within seed, then
mean$\pm$std over seeds. \\
\bottomrule
\end{tabular}
\caption{Complete final CDGP configuration. Settings are shared across all
categories and datasets unless a dataset-specific batch/input size is stated.}
\label{tab:reproduction_hyperparameters}
\end{table*}

\begin{table*}[ht!]
\centering
\footnotesize
\setlength{\tabcolsep}{2mm}
\begin{tabular}{@{}lp{0.69\textwidth}l@{}}
\toprule
Dataset & Partition and weak-label construction & Test overlap \\
\midrule
MVTec AD~2
& For each category and seed, pool the publicly labeled images, stratify by
the binary normal/defective label, and assign $80\%/20\%$ to train/test.
Split the resulting training partition again $80\%/20\%$ for fit/calibration
using a fixed stratified partition for each seed. Remove mask paths before
either training stage.
& 0 images \\
KSDD2
& Official train/test split; reduce each training mask once to a binary image
label; split official training images $80\%/20\%$ for fit/calibration.
& 0 images \\
VisA
& Official \texttt{2cls\_highshot} train/test CSV; use its normal/anomaly
training labels and split official training images $80\%/20\%$ for
fit/calibration.
& 0 images \\
\bottomrule
\end{tabular}
\caption{Partition provenance. Absolute resolved paths are hashed in released
seed/category manifests; all 24 MVTec AD~2 calibration--test audits report zero
overlap. Pixel masks remain inaccessible until final metric computation.}
\label{tab:split_provenance}
\end{table*}